\documentclass[12pt]{article}
\usepackage{graphicx}
\usepackage{natbib} 
\usepackage{url} 
 
\usepackage{hyperref}
\usepackage{graphicx}
\usepackage{makecell}
\usepackage{bm}
\usepackage{amssymb}
\usepackage{multirow}
\usepackage{arydshln}
\usepackage{xurl}
\usepackage[table]{xcolor}
\definecolor{light-gray}{gray}{0.9}
\usepackage{cancel}
\usepackage{color, colortbl}
\usepackage{amsfonts} 
\usepackage{arydshln}
\usepackage{amssymb}
\usepackage{bbm}

\RequirePackage{amsthm,amsmath}

\usepackage{breqn}

\usepackage{algorithmic}

\usepackage{amsmath}
\usepackage{array}
\usepackage{makecell}
\usepackage{mathtools}
\usepackage{xpatch}

\newcommand{\blind}{0}

\date{}

\begin{document}

\def\spacingset#1{\renewcommand{\baselinestretch}%
{#1}\small\normalsize} \spacingset{1}


\if0\blind
{
  \title{\bf DeepMIDE: A {Multi-Output} Spatio-Temporal Method for Ultra-Scale Offshore Wind Energy Forecasting}
  \author{
Feng Ye$^1$, Xinxi Zhang$^2$, Michael L. Stein$^3$, Ahmed Aziz Ezzat$^{4,5,*}$ \vspace{0.25cm}\\
$^1$Department of Industrial Engineering, Clemson University \\
$^2$Department of Computer Science, Rutgers University\\
$^3$Department of Statistics, Rutgers University \\ 
$^4$Department of Industrial \& Systems Engineering, Rutgers University\\
$^5$University College, Korea University
  \\ $^*${Correspondence: aziz.ezzat@rutgers.edu}}
  \maketitle
} 
\fi

\if1\blind
{
  \bigskip
  \bigskip
  \bigskip
  \begin{center}
    {\LARGE\bf Title}
\end{center}
  \medskip
} \fi

\bigskip

\begin{abstract}
To unlock access to stronger winds, the offshore wind industry is advancing {towards} significantly larger and taller wind turbines. This massive upscaling motivates a departure from wind forecasting methods that traditionally focused on a single representative height. To fill this gap, we propose DeepMIDE\textemdash a statistical deep learning method which jointly models the offshore wind speeds across space, time, and height. DeepMIDE is formulated as a {multi-output} integro-difference equation model with a multivariate nonstationary kernel characterized by a set of advection vectors that encode the physics of wind field formation and propagation. Embedded within DeepMIDE, an advanced deep learning architecture learns these advection vectors from high-dimensional streams of exogenous weather information, which, along with other parameters, are plugged back into the statistical model for probabilistic {multi-height} space-time forecasting. 
Tested on real-world data from offshore wind energy {areas} in the Northeastern United States, the wind speed and power forecasts from DeepMIDE are shown to outperform those from prevalent time series, spatio-temporal, and deep learning methods.
\end{abstract}

\noindent%
{\it Keywords:}  Forecasting, 
Integro-Difference Equation, 
Statistical Deep Learning, Spatio-Temporal Modeling, Wind Energy 

\spacingset{1.8} 
\section{Introduction}\label{intro}
Offshore wind energy is poised to play a major role in the worldwide transition towards cleaner and more sustainable energy systems \citep{musial2023offshore}. 
To harness stronger winds, the offshore wind industry is advancing with the next-generation offshore wind turbines, which are going to be significantly taller and larger than conventional wind turbines. With hub heights exceeding $140$ meters (m) and rotor diameters as large as $220$m, these ultra-scale turbines are set to become the largest rotating machines on earth, approaching the altitude of many of the world's tallest landmarks \citep{ gaertner2020definition, GE, 22mwturbine}. By producing significantly higher per-capita energy output using the same wind resources, considerable reductions in the cost of harnessing wind energy can be realized relative to conventional wind turbine designs that typically have had hub heights and rotor diameters of $100$m or less \citep{doefuture, shields2021impacts}. 

While the massive upscaling of offshore wind turbines unlocks significant benefits related to the ``economy of scale,'' it also raises considerable challenges pertaining to the optimal operation of those massive-scale assets in a highly dynamic and uncertain environment. Today, the reliable management of offshore wind farms heavily relies on short-term wind speed and power forecasts to mitigate the uncertainty of the intermittent wind resources \citep{giebel2016wind}. Those forecasts have been traditionally issued using statistical models that describe the temporal evolution of wind conditions at a single representative height, potentially combining information from multiple geographical locations \citep{sweeney2020future}. By doing so, those models compress the three-dimensional reality of the wind field (space $\times$ time $\times$ height) into a simpler, two-dimensional ``world view'' (space $\times$ time) which completely overlooks the vertical dimension of the wind field. {In the context of this paper}, ``space'' refers specifically to the horizontal dimensions (latitude and longitude), while “height” refers to the vertical dimension (altitude). 

While this simplification may be tenable for conventional wind turbines, it is indeed not plausible for the next-generation of offshore wind turbines\textemdash there is growing evidence in the wind power engineering literature that larger-scale wind turbines experience a significant influence from vertical wind shear on the turbine's power production. Hence, the power output of those ultra-scale generators more accurately correlates with the full vertical wind profile swept by the now-much-larger rotor diameter, rather than with a single horizontal slice representing the average hub-height wind speeds
\citep{antoniou2009wind,wagner2011accounting, barthelmie2020power}. This observation motivates us to seek more sophisticated models for offshore wind forecasting in which the evolution of wind conditions at multiple locations \textit{and} heights are described in tandem. Our survey of the literature, presented in Section \ref{review}, reveals that there is very limited work on multi-height space-time wind forecasting. We believe that, until recently, this has been largely due to the lack of a practical need for such models. Evidently, this is no longer the case with the emergence of ultra-scale wind turbines. 

As we argue in this paper, making the leap from {single- to} {multi-height} space-time wind forecasting is not a trivial task, primarily due to the complex and non-separable dependencies of the full-fledged wind field across the space-time-height trio. To fill this gap, we propose a {multi-output} spatio-temporal model with a multivariate, {nonstationary 
kernel,} which\textemdash unlike existing space-time wind forecasting approaches\textemdash is capable of adequately modeling the complex dependencies in local wind speeds across multiple heights. Embedded within the statistical model, an advanced deep learning architecture fully exploits the wealth of exogenous information available to the forecaster (typically in the form of high-dimensional image streams) in order to learn and update key physically meaningful kernel parameters that encode physically relevant information about wind field dynamics. In that way, our proposed approach retains the desirable properties of a statistical model, while embedding a deep learning architecture to act as an internalized ``physics extractor'' that exploits the large amounts of high-dimensional exogenous information that is difficult to process using standard statistical machinery. 
Experiments using real-world {offshore wind} data from 
{the Northeastern United States\textemdash a region with significant offshore wind energy potential\textemdash demonstrate} that the wind speed and power forecasts from the proposed model are of considerably higher quality than those from prevalent time series, deep learning, and space-time methods. 

The remainder of this paper is organized as follows. Section \ref{review} reviews the literature on space-time forecasting {for} wind energy applications. Section \ref{data} presents the real-world data used in this study and its relevance to the {planned offshore wind energy developments} in the Northeastern United States. In Section \ref{method}, we present the building blocks of the proposed model, followed by Section \ref{results} where forecast evaluations, results, and findings are discussed. Section \ref{conclusion} concludes the paper and highlights future research directions.



\section{Literature review}\label{review}
The engineering literature on wind turbine modeling acknowledges the influence of 
vertical wind speed variations on the performance of large-scale wind turbines \citep{antoniou2009wind,wagner2011accounting, lee2015power, barthelmie2020power, iec2022}. 
This literature primarily focuses on modeling a turbine's power output, conditional on a fully known wind speed profile. 

In practice, however, wind speed\textemdash the key determinant of wind power output\textemdash is the major source of uncertainty in wind energy harnessing, and is not known to wind farm operators in advance. Thus, to optimally operate an (offshore) wind farm, accurate wind speed and power forecasts are needed at various forecast horizons, ranging from a few minutes up to several days ahead. Those forecasts directly inform critical decisions made by energy producers and grid operators alike, including power production estimation \citep{zhu2021multi, nasery2023yaw}, electricity market participation and trading \citep{Pinson2013}, operations and maintenance scheduling \citep{petros2021, papadopoulos2023joint}, economic dispatch and unit commitment \citep{xie2013short,barry2022risk}, among others. Despite the general consensus in the wind engineering community on the importance of considering the vertical wind profile, there are very few studies in the forecasting literature on multi-height wind forecasting \citep{lin2020wind, saxena2021offshore}. Even these studies are only focused on a single location, thereby ignoring the spatial variations and correlations in the wind field. 

On the other hand, there is an extensive body of literature on space-time wind forecasting, which focuses on modeling the spatial and temporal variations and correlations, but ignores the vertical dimension of wind fields.
{A relevant class of models therein are geostatistical approaches, such as spatio-temporal Gaussian processes (GPs).}
GPs make space-time forecasts that are enabled by a covariance function describing the dependence structure over space and time \citep{GneitingCov,SteinCov,AzizGP}. A central challenge in GP-based models is to propose a mathematically permissible covariance function that adequately models the complex dependencies in wind fields. {An} important aspect of that complexity is to model the effect of time-varying wind advection, which gives rise to an asymmetry in the spatial-temporal correlations. This complexity requires a covariance model that relaxes the convenient (but unrealistic) assumptions of stationarity, separability, and full space-time symmetry \citep{gneiting2006geostatistical,AzizRS}. 
{For example, Lagrangian covariance functions introduce advection-specific parameters that can explicitly encode the effect of wind propagation on the magnitude and asymmetry of the resulting space-time correlations \citep{cox1988simple, ma2003families}.} 

An alternative paradigm to geostatistical approaches are the so-called \textit{dynamic spatio-temporal models (DSTMs)} which describe the conditional evolution of a spatial process over time \citep{wikle2010general}. {In this context, a} prevalent example is the integro-difference equation (IDE) framework, which models the dependence between a spatial process at a future time conditioned on the same process at the present time through an integral operator with a spatial redistribution kernel \citep{wikle1999dimension}. IDE models have been proposed to predict spatio-temporal cloud movement \citep{Wikle2002}, precipitation \citep{xu2005, liu2022statistical}, aerosols \citep{calder2011modeling}, and more recently, wind power \citep{ye2024ieee}. {Both geostatistical models and DSTMs offer distinct advantages: geostatistical approaches are typically regarded as capable surrogate models for complex processes, whereas DSTMs can be highly effective for extrapolating dynamic processes by virtue of their explicit conditional evolution mechanism.}



Different in principle than the first two statistical approaches (i.e., geostatistical models and DSTMs), \textit{deep learning methods} have demonstrated significant promise in wind forecasting applications by developing architectures that are tailored to extract complex patterns from spatio-temporal wind data \citep{dlreview4wind}. {These include} models based on convolutional neural networks (CNNs) \citep{zhu2019learning},  recurrent neural networks (RNNs) \citep{ghaderi2017}, graph neural networks (GNNs) \citep{GNNwind1, GNNwind2}, and more recently, transformer models \citep{bentsen2023spatio}. Despite their demonstrated predictive performance, deep learning {methods} still fall short in terms of uncertainty quantification and interpretability. Thus, 
recent lines of research seek hybrid methods, referred to collectively as \textit{statistical deep learning approaches}, in which elements or characteristics of a deep learning architecture are embedded within a statistical model to perform certain learning tasks. For example, deep Kalman filters \citep{krishnan2015deep} and deep state-space models \citep{deepssm} learn the parameters of a state-space model using multi-layer perceptrons (MLPs), and recurrent neural networks (RNNs), respectively. Deep autoregressive models (DeepAR) entail a sequence of RNNs that are structured in an autoregressive-inspired manner \citep{deepar}, wheras deep IDE models invoke a CNN to learn the IDE redistribution kernel parameters using past process observations \citep{de2019deep,zammit2020deep}. 

Departing from the existing literature, we propose a {multi-output} IDE model characterized by a multivariate, {nonstationary 
redistribution kernel with functional parameters, which} is tailored to capture the complex dependencies and interactions of local wind fields across the space-time-height trio. To exploit the wealth of exogenous information available to the forecaster (typically in the form of high-dimensional images), we embed an advanced deep learning architecture within the {multi-output} IDE model in order to learn key physically meaningful kernel parameters that encode physically relevant information about local wind field dynamics. The proposed model is therefore statistical in nature, enabling full probabilistic inference and forecasting across space, time, and height, with the internalized deep learning architecture acting as a ``physics extractor,'' to learn a low-dimensional set of physically relevant parameters using high-dimensional data 
 that are difficult to process using standard statistical machinery. 

Our proposed model departs from the vast majority of space-time wind forecasting models (be it geostatistical, DSTMs, or deep-learning-based) that adopt a two-dimensional world view of the wind field (space $\times$ time) by focusing on a single representative height. It belongs to the class of statistical deep learning models reviewed above, and especially builds on the deep IDE models therein \citep{de2019deep, zammit2020deep}, but differs in two key aspects. First, it generalizes the {single-output} IDE approaches to the multivariate setting to model the complex and time-varying dependencies across space, time, and height. In that way, the proposed model is regarded as {a multi-output IDE},
akin to how multi-output GPs generalize their single-output counterparts. {The need for the proposed modeling framework is motivated by our data analysis, presented in Section \ref{data}, where the multivariate time series representing the multi-height offshore wind speeds are shown to exhibit complex, time-varying cross-correlations, calling for a multi-output dynamical approach 
in which the height dimension is modeled as a physically meaningful vertical axis rather than merely another spatial coordinate. This multivariate time series approach enables us to learn the complex cross-height dependencies and interactions that cannot be handled by simply treating height as a geospatial coordinate. Another relevant work to this discussion is the multivariate spatio-temporal mixed effects model (MSTM) introduced by \cite{mstm}, which is designed to model correlated multivariate areal data over space and time. While the MSTM falls under the broader class of multivariate DSTMs, it is not based on the IDE framework, and hence is not designed to model the complex dependencies inherent in the full-fledged wind field across space, time, and height.}

Second, unlike existing deep IDE models which use information that is endogenous to the process (e.g., past observations) to train simple deep learning architectures (typically a CNN), our model exploits the wealth of exogenous information available to the forecaster, typically in the form of high-dimensional image streams. For example, when forecasting wind speeds, one has access to large amounts of high-dimensional data from regional numerical weather models about the meso-scale wind velocities, pressure and temperature systems, etc. that encode rich physical information about the local wind dynamics. To fully unlock access to this exogenous information, we propose to embed a transformer model\textemdash an advanced deep learning architecture with evidenced success in computer vision and high-dimensional data mining \citep{transformer}\textemdash within the {proposed multi-output} IDE framework as a ``physics extractor,'' i.e., to learn and update a low-dimensional set of time- and height-varying parameters encoding the physics of local wind field formation and propagation. {This  not only enables tapping into exogenous, high-dimensional data streams for improved forecasting (instead of being confined to endogenous, historical time series as in previous deep IDE models), but further elevates the role of the embedded deep learning architecture from an opaque black-box to a ``physics extractor'' by learning key physically meaningful parameters that directly inform the statistically derived forecast.}

\section{Data description and analysis}\label{data}
This work is motivated by the 
{potential} large-scale offshore wind  energy developments in the Northeastern United States, and in particular, the NY/NJ Bight \citep{leasenj}. 
{The future offshore wind energy areas in this region, as well as in other locations world-wide, are expected to install ultra-scale offshore wind turbines (hub heights $> 140$m and rotor diameters $> 200$m) \citep{GE}}. We make use of two sources of data with varying spatial and temporal resolutions: (\textit{i}) A set of wind speed measurements collected, at multiple heights, by three floating lidar buoys; and (\textit{ii}) a set of numerical weather predictions (NWPs), of multiple meteorological variables, obtained from a regional numerical weather model. Details of both sets of data are described next. 
\begin{figure}[htbp!]
\centering  
\includegraphics[trim = 0cm .75cm 0cm 0cm, width = \linewidth]{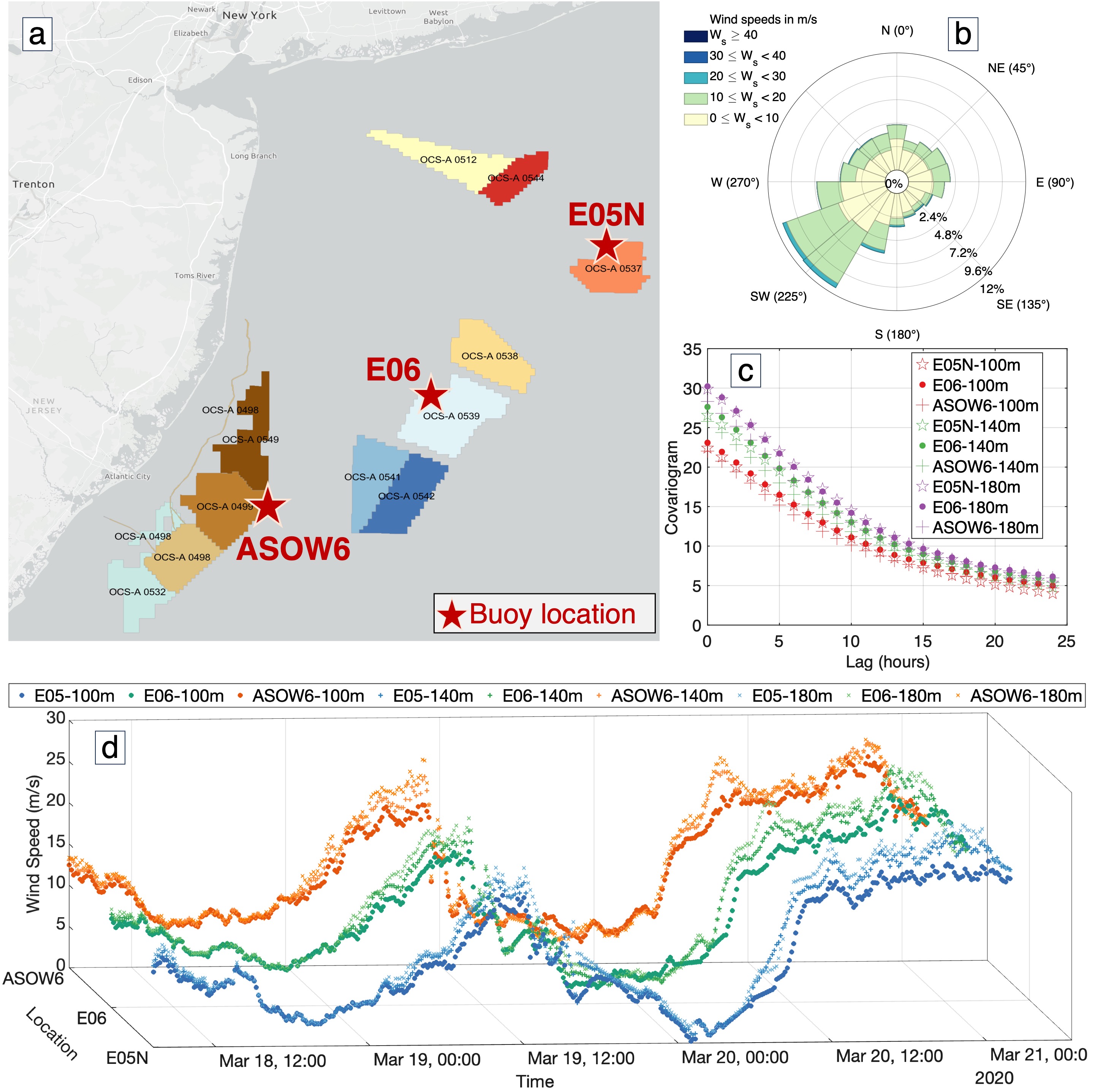}
\caption{(a) Locations of the three sites, E05N, E06, and ASOW6, on top of the offshore wind energy areas in the NY/NJ Bight. The background map is generated using the Northeast Ocean Data Portal \citep{map}; (b) Wind rose plot for the wind speeds, averaged over space and height; (c) {The empirical temporal covariogram of wind speeds at multiple heights (different colors) and locations (different markers);} {(d) Time series of wind speeds at all locations (different colors) and heights (different markers), showing strong spatial, temporal, and vertical dependencies.} 
}
\label{figlocation}  
\end{figure}  

\subsection{Local observations from the NY/NJ Bight}\label{truedata}
Wind speeds are measured using three floating lidar buoys, namely E05 Hudson North (E05N), E06 Hudson South (E06) \citep{measurement}, and ASOW6 \citep{asow6}. Figure \ref{figlocation}(a) shows the locations of the three sites on top of the designated offshore wind energy areas in the NY/NJ Bight. E05N is the farthest location from shore, and is $\sim$\hspace{-0.05mm}$77$ kilometers (km) away from E06. ASOW6 is the southernmost and closest location to shore. E06 and ASOW6 are $\sim$\hspace{-0.05mm}$54$ km apart. 
The observations are recorded in $10$-min resolution, for a total duration of approximately $8.5$ months spanning the years of 2020 and 2021, at three heights: $100$m, $140$m, and $180$m. The data from ASOW6 had {extended periods of missing observations,} 
which were imputed using another buoy in close proximity. 
Figure \ref{figlocation}(b) {shows} the wind rose plot of the data (pooled across all heights and locations), {suggesting} dominant southwesterly winds. 

{
Figure \ref{figlocation}(c) shows the empirical temporal covariograms (commonly referred to as autocovariance functions) for the offshore wind speeds across all locations and heights. 
Some clear temporal correlation patterns are observed: (\textit{i}) For all nine time series, the empirical covariograms show strong correlations at short time lags that decrease with lag but are still distinctly positive at the 24-hour lag; (\textit{ii}) For different locations (E05N, E06, ASOW6) at the same height, the temporal correlation structure appears to be similar, but with slightly lower overall variability at ASOW6, the site nearest to the coast; (\textit{iii}) Comparing different heights (100m, 140m, 180m), the covariogram indicates consistently higher variability at greater altitudes (180m) compared to lower altitudes (140m and 100m).}
{Figure \ref{figlocation}(d) shows the multivariate time series of the wind speeds at all locations and heights during a select period in March 2020, clearly demonstrating strong spatial, temporal, and vertical dependencies, further motivating the need for a joint modeling approach across the space-time-height trio, as the one advocated in this paper.}


\subsection{Exogenous information from a regional NWP model }\label{ru-wrf}
\begin{figure}[h]  
\centering  
\includegraphics[trim = 0cm .75cm 0cm 0cm, width = \linewidth]{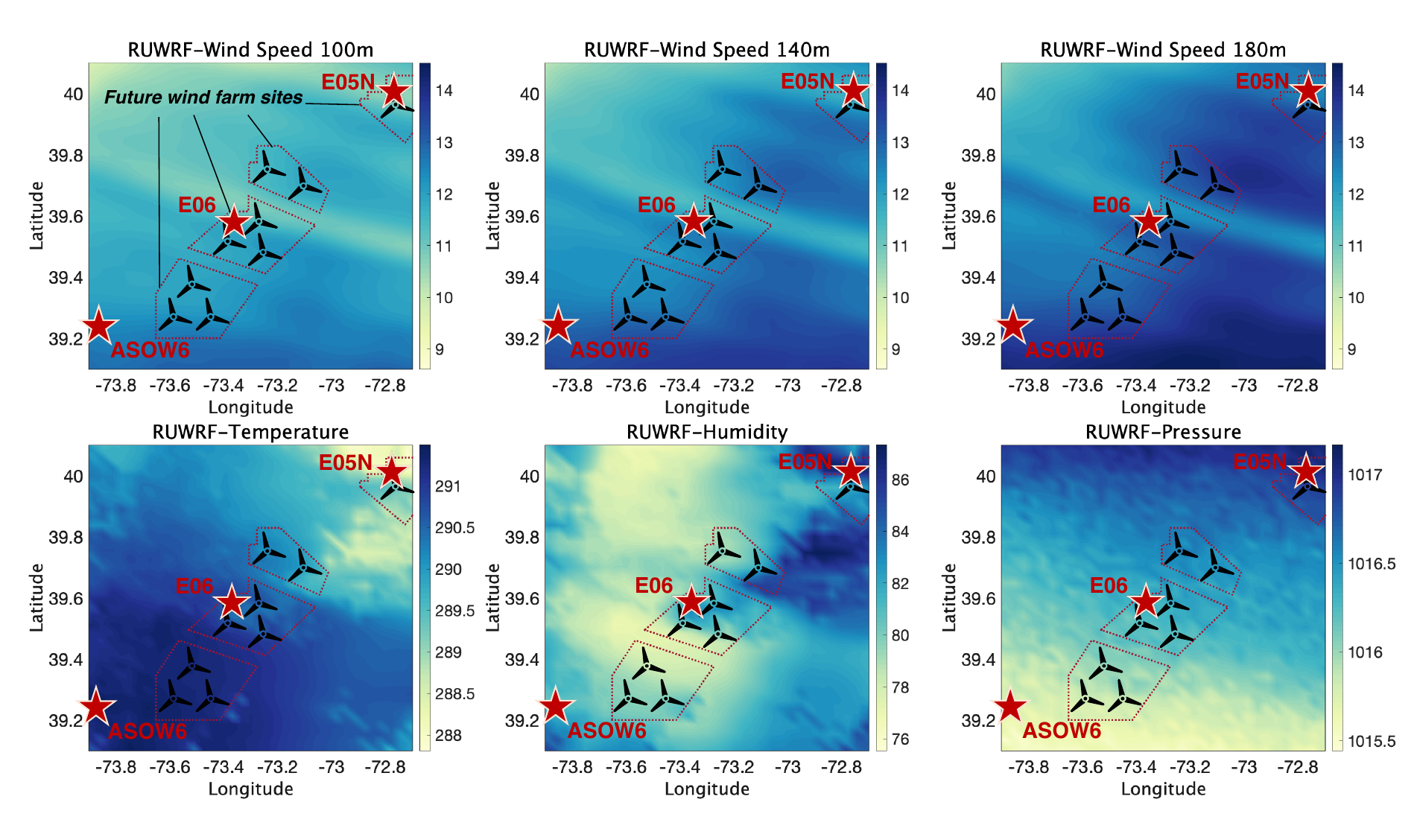}
\caption{Spatial weather maps from RU-WRF (wind speed at 100/140/180m, sea-surface temperature, humidity, and pressure) on June 11th, 2021 at 8:00 GMT on top of the {planned} offshore wind energy areas (denoted by dashed polygons) in the NY/NJ Bight. Red stars denote the spatial locations where local measurements are available.}
\label{ruwrf}  
\end{figure}

A daily real-time version of the Weather Research and Forecasting (WRF) model, called RU-WRF, generates 
hourly outputs, at $3$-km resolution, for several meteorological variables, including multi-height wind speeds, sea-surface temperature, humidity, and pressure \citep{dicopoulos2021weather}. 
The model is tailored to the U.S. Mid- and North- Atlantic offshore wind energy areas.  It has been independently validated by the National Renewable Energy Laboratory (NREL) \citep{optis2020validation} and has been continuously evaluated and improved since then \citep{dicopoulos2021weather}. Its spatial domain spans the coastal region from southern Massachusetts to North Carolina. 
Figure \ref{ruwrf} shows RU-WRF outputs, in the form of spatial weather maps, on a select day and time in June, 2021 on top of the offshore wind energy areas in the NY/NJ Bight. 


\subsection{Asymmetry analysis}\label{asy}
Due to advection effects, wind fields {can} exhibit strong signs of asymmetry in spatio-temporal dependencies. This means that along-wind dependence (i.e., the dependence between the wind conditions at an upstream location at time $t$ and that at a downstream location at time $t + u$, {where $u>0$ is a time lag}) is expected to be stronger than opposite-wind dependence  (i.e., the dependence between the wind conditions at a downstream location at time $t$ and that at an upstream location at time $t + u$) \citep{GneitingCov,SteinCov, AzizIEEE}. Thus, adequate models should be able to capture those complex interaction effects. To demonstrate the existence of asymmetry in our data, we first de-trend the time series data at each height and location by fitting a time- and height-specific trend comprising a series of diurnal harmonics, as expressed in (\ref{eq:diurnal}). 
\begin{equation}
\small
D_t^{(g)}(\mathbf{s})=d_0^{(g)}(\mathbf{s})+d_1^{(g)}(\mathbf{s}) \sin \left(\frac{2 \pi t}{24}\right) +d_2^{(g)}(\mathbf{s}) \cos \left(\frac{2 \pi t}{24}\right) +d_3^{(g)}(\mathbf{s}) \sin \left(\frac{4 \pi t}{24}\right)+d_4^{(g)}(\mathbf{s}) \cos \left(\frac{4 \pi t}{24}\right),
\label{eq:diurnal}
\end{equation}
{where $D_t^{(g)}(\mathbf{s})$ is the diurnal trend at time $t \in \mathbb{Z}^+$,  spatial location $\mathbf{s}\in\mathbb{R}^2$ (longitude, latitude), and height $ g \in \mathbb{R}^+ $, whereas $d_0^{(g)}(\mathbf{s}), \hdots, d_{4}^{(g)}(\mathbf{s})$ are the corresponding regression coefficients, which are location- and height-dependent.} The de-trended data is then used to compute an estimate of asymmetry at the $g$th height, denoted as $a^{(g)}(\mathbf{s}_i, \mathbf{s}_j, u)$, and defined as the difference in spatio-temporal semi-variograms (as a proxy for differences in spatio-temporal covariances). This is expressed in (\ref{eq:asy}).
\begin{equation}
  a^{(g)}(\mathbf{s}_i, \mathbf{s}_{j}, u) := \delta^{(g)}(\mathbf{s}_i, \mathbf{s}_{j}, u) - \delta^{(g)}(\mathbf{s}_{j}, \mathbf{s}_{i}, u),
  \label{eq:asy}
\end{equation}
where $\mathbf{s}_i$ and $\mathbf{s}_{j}$ denote the coordinates of the $i$th and $j$th locations, and $\delta^{(g)}(\cdot,\cdot,\cdot)$ is the spatio-temporal semi-variogram at height $g$, which is defined as in (\ref{eq:vario}).
\begin{equation} \label{eq:vario}
\delta^{(g)}\left(\mathbf{s}_{i}, \mathbf{s}_{j}, u\right)=\frac{1}{2(N-u-1)} \sum_{t=1}^{N-u-1}\left\{r^{(g)}_{t+u}\left(\mathbf{s}_i\right)-r^{(g)}_t\left(\mathbf{s}_{j}\right)\right\}^{2},
\end{equation}
where $N$ is the total number of data points, and  $r^{(g)}_{t}\left(\mathbf{s}_i\right)$ denotes the value of the de-trended time series at location $\mathbf{s}_i$, time $t$, and height $g${, that is, $r^{(g)}_{t}\left(\mathbf{s}_i\right) =  Z^{(g)}_t(\mathbf{s}_i) - D_t^{(g)}(\mathbf{s}_i)$, such that $Z^{(g)}_t(\mathbf{s}_i)$ is the wind speed observation at location $\mathbf{s}_i$, time $t$, and height $g$.}

\begin{figure}[h]  
\centering  
\includegraphics[trim = 0cm .75cm 0cm 0cm, width = \linewidth]{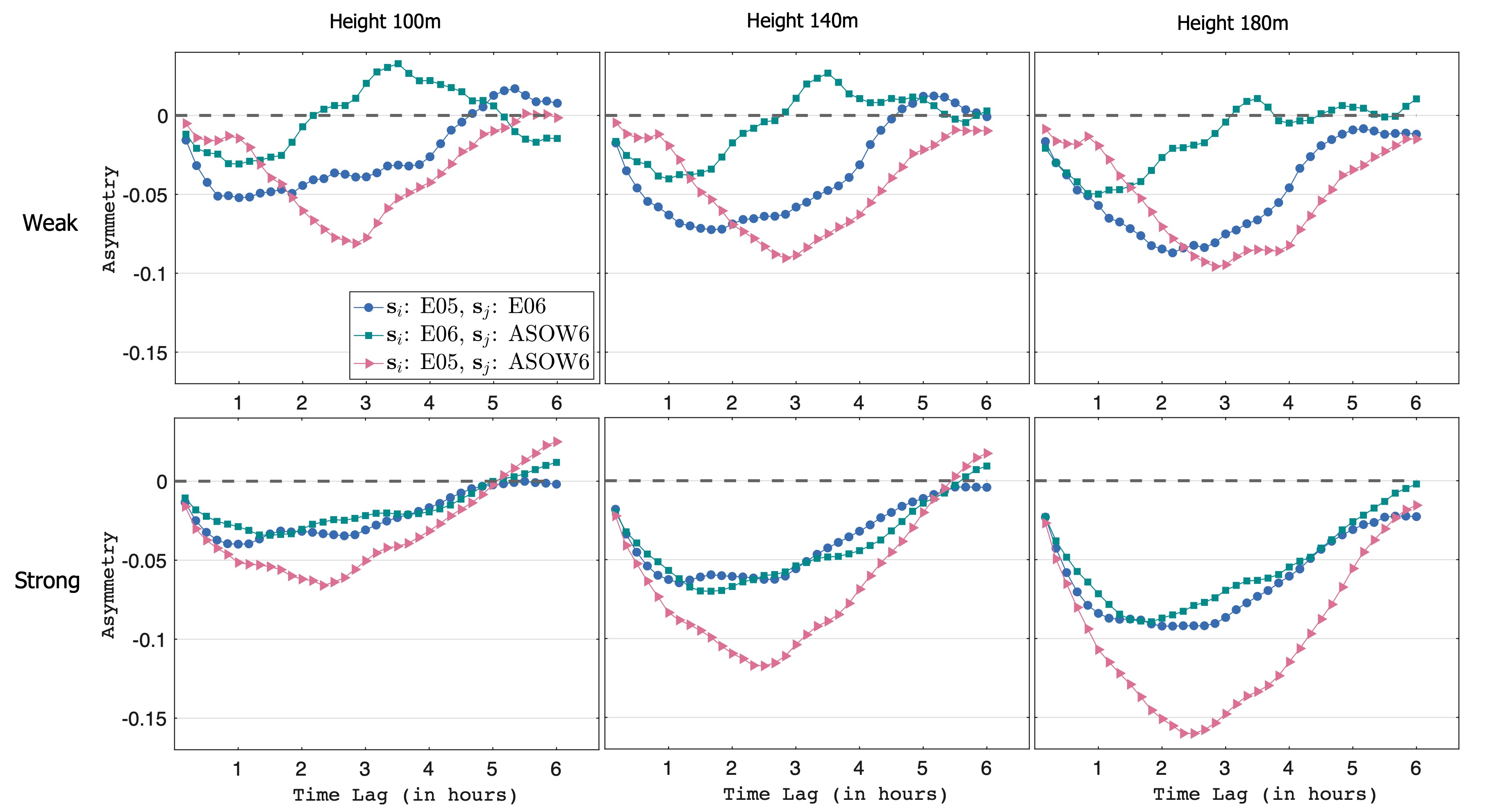}
\caption{{Asymmetry in space-time dependencies, versus the time lag in hours, for different heights (different columns). The top and bottom rows represent weak and strong wind regimes, respectively. Different colors denote different pairs of locations.}}
\label{fig:asy}  
\end{figure}

Figure \ref{fig:asy} shows the average values of $a^{(g)}(\cdot, \cdot, \cdot)$ versus the time lag $u$ (x-axis), for different heights (different columns), partitioned into two wind speed regimes (different rows), representing weak and strong winds respectively (as defined by a threshold of $8$ m/s). 
A few key observations can be deduced. First, it is clear that there is noticeable levels of spatial-temporal asymmetry in our data, as evident by the departure from the full-symmetry baseline represented by the horizontal dashed line at 0. This asymmetry is the result of the wind propagation across the prevailing southwesterly wind (recall Figure \ref{figlocation}(b)). The asymmetry appears to be maximal at the $\sim$\hspace{-.02mm}$1$-$3$-hour time lag, which aligns with the expected time for wind conditions to propagate across the spatial locations considered. Interestingly, asymmetry appears to be stronger with larger altitudes and stronger winds. 
This means that changes in wind speed over time and height would be accompanied by changes in the strength of asymmetry. An adequate model would not only need to account for the existence of asymmetry, but also for the time- and height-varying dependence structure driven by wind advection dynamics. Existing asymmetric space-time models typically assume a constant level of asymmetry across height and time \citep{SteinCov, gneiting2006geostatistical}, or at best, may model some time-varying advection effects, but overlook the vertical variations in asymmetry levels \citep{ye2023airu, ye2025improved}. {The above observations warrant a multi-output dynamical forecasting method that can adequately model the complex, time-varying dependence structure found in the multivariate time series data representing the offshore wind speeds across space, time, and height. This will be the focus of Section \ref{method}}.

\section{Metholodogy}\label{method}
To model the complex dependencies in the local wind fields across space, time, and height, we propose a {multi-output} deep IDE model with a multivariate, {nonstationary kernel,} 
for which key physically meaningful parameters are learned using an embedded deep learning model. Our approach is dubbed as the {Deep} {{M}ulti-output} {I}ntegro-{D}ifference {E}quation model, or in short, DeepMIDE. We start by a brief background to the IDE model, then present the essence and technical details of DeepMIDE. 


\subsection{Background to the IDE framework}\label{sec:ide}
Stochastic IDE models are a class of hierarchical space-time models comprised of a data model and a process model \citep{cressie2011statistics}. The data model relates a latent state process to a set of observable measurements. In the linear-Gaussian case, it can be written as in (\ref{eq:datamodel_ide}). 
\begin{equation}
\mathbf{z}_t = \mathbf{h}_t\mathbf{y}_t+\boldsymbol{\xi}_t,
\label{eq:datamodel_ide}
\end{equation}
where $\mathbf{z}_t$ is an $m$-dimensional vector of observations at time $t$ {and $m$ spatial locations, such that $\mathbf{z}_t \equiv (z_t(\mathbf{s}_1), \hdots, z_t(\mathbf{s}_{m}))^\top$}, 
while $\mathbf{y}_t$ is the correspondent $n$-dimensional vector of the latent state process at time $t$, {i.e., $\mathbf{y}_t\equiv (y_t(\mathbf{s}_1^*), \hdots, y_t(\mathbf{s}_{n}^*))^\top$}. 
The matrix $\mathbf{h}_t$ is an $m\times n$ observation mapping matrix, whereas $\boldsymbol{\xi}_t$ is an $m$-dimensional vector of residuals, which are independent of $\mathbf{y}_t$. 
Assuming locally linear dynamics, the process model describes the conditional first-order evolution of the state process through a spatial integral operator, as expressed in (\ref{eq:processmodel_ide}). 
\begin{equation}
{y}_t(\mathbf{s})=\int_{\mathcal{D}_s} k(\mathbf{s}, \mathbf{x}; \boldsymbol{\theta}) {y}_{t-1}(\mathbf{x}) d \mathbf{x}+\nu_t(\mathbf{s}), \quad \mathbf{s} \in \mathcal{D}_s,
\label{eq:processmodel_ide}
\end{equation}
where ${y}_t(\cdot)$ denotes the spatial process {across a region of interest $\mathcal{D}_s\in \mathbb{R}^2$} at time $t$. The error process $\nu_t(\cdot)$ is independent in time but may be correlated in space, whereas $k(\mathbf{s}, \mathbf{x} ; \boldsymbol{\theta})$ is a spatial redistribution kernel characterized by a set of parameters denoted by $\boldsymbol{\theta}$. 
{Discretizing the integral in (\ref{eq:processmodel_ide}) at $n$ spatial locations yields the following latent process model: 
\begin{equation}
{
{y}_t(\mathbf{s})=\sum_{\mathbf{x}=\mathbf{s}_1^*}^{\mathbf{s}_n^*} \tilde{k}(\mathbf{s}, \mathbf{x}; \boldsymbol{\theta}) {y}_{t-1}(\mathbf{x})+\nu_t(\mathbf{s}),}
\label{eq:processmodel_ide2}
\end{equation}
where $\tilde{k}(\mathbf{s}, \mathbf{x}; \boldsymbol{\theta})$  is the correspondent redistribution kernel of the discretized process model.}

Stochastic IDEs have received significant attention in the space-time modeling literature due to their versatility and ability to integrate mechanistic descriptions of physical and {scientific} knowledge. To date, the overwhelming majority of IDE models have been {single-output}, i.e. they model a single stochastic process over space and time \citep{Wikle2002, wikle2005, xu2005, wikle2011, calder2011modeling, liu2022statistical, ye2024ieee}. However, many real-world processes do not exist in isolation; they are influenced by, and interact with, other concomitant processes. Very little attention has been dedicated to {multi-output} IDE models, and even less so for wind forecasting applications. {In response, Section \ref{sec:mide} presents} a {multi-output} IDE to model and forecast the wind energy resource over space, time, and height, as motivated by {the emergence of ultra-}scale offshore wind turbines. 

\subsection{A {multi-output} IDE model for multi-height space-time offshore wind forecasting}\label{sec:mide}
{For notational consistency, we will use uppercase notation to distinguish between the single-output setting presented in Section \ref{sec:ide} and the multi-output setting, to be discussed below.} Let $Z^{(g)}_t(\mathbf{s})$ be the wind speed observation at location $\mathbf{s}$, time $t$, and height $g$. 
We denote by $\mathbf{Z}^{(g)}_t \equiv (Z^{(g)}_t(\mathbf{s}_1), \hdots, Z^{(g)}_t(\mathbf{s}_{m}))^\top$ the $m$-dimensional spatial vector of observations at a particular time $t$ and height $g$, whereas $\mathbf{Z}_t \equiv (\mathbf{Z}^{(1)}_t, ..., \mathbf{Z}^{(G)}_t)^\top$ is the $mG \times 1$ matrix comprising the observations at time $t$, {across the $m$ locations and $G$ heights 
where measurements are available.} 
We express the data model for the {multi-output} IDE as in (\ref{eq:data}).
\begin{equation}
\mathbf{Z}_t = \mathbf{H}_t\mathbf{Y}_t+\boldsymbol{\epsilon}_t,
\label{eq:data}
\end{equation}
where $\mathbf{Y}_t \equiv (\mathbf{Y}^{(1)}_t, \hdots, \mathbf{Y}^{(P)}_t)^\top$ is the $nP \times 1$ matrix representing the underlying state process at time $t$, {across $n$ locations and $P$ heights}. 
The matrix   
$\mathbf{H}_t$ is a $mG\times nP$ observation mapping matrix, 
whereas {$\boldsymbol{\epsilon}_t$ denotes the $mG \times 1$ vector of errors, 
which are assumed to follow a zero-mean multivariate Gaussian that is independent in time but not necessarily in space}, with a spatial covariance matrix $\Sigma_{\boldsymbol{\epsilon}}$, for which the entries are determined using a squared exponential covariance function 
$c_\epsilon(d) = \sigma_{\epsilon}^2\exp{(-\frac{1}{2}(\frac{\|d\|^2}{\ell^2_{\epsilon}})})$, where $d$ is the pairwise spatial distance, while $\sigma_\epsilon$ and $\ell_\epsilon$ are the variance and length-scale parameters, respectively. {The spatial dependence assumption on the error term can be dropped, but is found to slightly benefit the model adequacy by capturing residual spatial dependencies.} The Gaussian assumption is non-problematic {either} after an appropriate Box-Cox transformation {applied to the pooled data across all locations and heights}, which makes the wind speeds nearly Gaussian. 
For our data, the Box-Cox parameter is estimated at $0.504$, which is almost equivalent to the square root transformations common for modeling wind speeds in the literature \citep{haslett1989space,SteinCov, gneiting2006calibrated}. 
{The forecasts can be retrieved through an inverse Box-Cox transformation directly applied on the model's outputs. 
Since the inverse Box-Cox transformation yields the median of the forecast distribution, we followed the approximation described in \cite{hyndmanforecasting}, which recovers the distribution mean using a second-order Taylor series expansion. The details of the Box-Cox transformation and its inverse are described in the supplementary material document appended to this
manuscript.}

Considering {multi-output spatio-temporal processes representing the offshore wind speeds over space, time, and height}, 
$\{Y_t^{(p)}(\cdot)\}$, the {multi-output} version of the stochastic IDE process model can be written as in (\ref{eq:mide}).
\begin{equation}\label{eq:mide}
{Y_t^{(p)}(\mathbf{s})=\sum_{\mathbf{x}=\mathbf{s}_1^*}^{\mathbf{s}_n^*} \sum_{q=1}^P k^{(p q)}(\mathbf{s}, \mathbf{x} ; t, \boldsymbol{\theta}_t^{(p q)})  Y_{t-1}^{(q)}(\mathbf{x})+\eta_t^{(p)}(\mathbf{s}),}
\end{equation}
such that $k^{(p q)}(\mathbf{s}, \mathbf{x}; t, \boldsymbol{\theta}_t^{(p q)})$ is the redistribution kernel of the {multi-output} IDE, which determines the contribution of height $q$ at location $\mathbf{x}$ and time $t-1$ in describing the height $p$ at location $\mathbf{s}$ at the current time $t$. The redistribution kernel is assumed to depend on a set of time-varying parameters, denoted by $\boldsymbol{\theta}_t^{(pq)}$. The state error vector at time $t$, {across $P$ heights and $n$ locations}, is denoted by {$\boldsymbol{\eta}_t = (\boldsymbol{\eta}_t(\mathbf{s}_1^*), ..., \boldsymbol{\eta}_t(\mathbf{s}_n^*))^\top$} such that $\boldsymbol{\eta}_t(\mathbf{s}) \equiv (\eta_t^{(1)}(\mathbf{s}), ..., \eta_t^{(P)}(\mathbf{s}))^\top$ is the corresponding error vector at location $\mathbf{s}$ and time $t$, across all $P$ heights. 
The  error term is assumed to be a zero-mean {multivariate} Gaussian which is independent in time but not necessarily in space, with a spatial covariance matrix $\Sigma_{\boldsymbol{\eta}}$, for which the entries are determined using a squared exponential covariance function $c_\eta(d) = \sigma_{\eta}^2\exp{(-\frac{1}{2}(\frac{\|d\|^2}{\ell^2_{\eta}})})$, where $\sigma_\eta$ and $\ell_\eta$ are the variance and length-scale parameters, respectively.

Modeling the redistribution kernel $k^{(p q)}(\mathbf{s}, \mathbf{x}; t, \boldsymbol{\theta}_t^{(p q)})$ is {a critical} aspect of IDE models. The asymmetry analysis in Section \ref{asy} demonstrated the need for capturing time- and height-varying advection dynamics. Here, we follow a similar pursuit to the recent work of \cite{madv} who introduced asymmetric cross-covariance functions with multiple advections. We leverage their key theorem to propose a multivariate redistribution kernel, as expressed in (\ref{eq:spmk}), that is tailored to model the dependence structure in offshore wind fields, considering time- and height-varying advections. 
\begin{equation}\label{eq:spmk}
k^{(pq)}\left(\mathbf{s}, \mathbf{x} ; t, \boldsymbol{\theta}_t^{(p q)}\right)={\sigma_k^2}\exp \left(-\frac{\|\mathbf{s}-\mathbf{x}-\boldsymbol{\theta}_{t}^p t+\boldsymbol{\theta}_{t}^q (t+1)\|^2}{\ell_{\mathbbm{1}_{p=q}}^2}\right),
\end{equation}
where {$\sigma_k$ is the gain parameter}, $\ell_{\mathbbm{1}_{(p=q)}}$ $\forall$ $p, q \in \{1, \hdots, P\}$ denote a set of height-dependent diffusion parameters, $\mathbbm{1}_{(\cdot)}$ is the indicator function, {whereas the spatial locations $\mathbf{s}$ and $\mathbf{x}$ are vectors in $\mathbb{R}^2$ comprising the latitude and longitude coordinates}.  
The kernel $k^{(pq)}\left(\mathbf{s}, \mathbf{x}; t, \boldsymbol{\theta}_t^{(p q)}\right)$ is characterized by a set of time- and height-varying advection vectors $\boldsymbol{\theta}_{t}^{(p q)} = \{\boldsymbol{\theta}_{t}^p, \boldsymbol{\theta}_{t}^q\}$, which collectively encode the physics of local wind field formation and propagation, and are assumed to be random in $\mathbb{R}^2$. 



Collecting all the advection vectors across {$P$} heights yields the set $\boldsymbol{\theta}_t = \{\boldsymbol{\theta}_{t}^{p}\}_{p=1}^P$. Doing so across all time indices yields the parameter set $\boldsymbol{\Theta}=\{\boldsymbol{\theta}_t\}_{t=1}^{T}$. This approach, which treats the advection parameters as time- and height-varying, is in line with  
the {complex dependencies observed in multi-height offshore wind speeds, as shown in Section \ref{asy}}. 
However, it requires an estimation of a very large number of parameters at each time $t$ and height $p$. 
Instead, we propose to embed a mapping function, denoted by $\psi(\cdot)$, to represent the parameter set $\boldsymbol{\Theta}$, given streams of exogenous weather maps 
$\mathcal{X}=\{\mathbf{X}_t\}_{t=1}^T$ from a regional numerical model, where $\mathbf{X}_t \in \mathbb{R}^{C \times \mathcal{W}}$ denotes a set of weather maps at time $t$ representing $C$ exogenous variables, with each map acquired on a bounded rectangle {in} $\mathbb{R}^2$ denoted by $\mathcal{W}$. An example of $\mathbf{X}_t$ in the context of our problem are the weather maps shown in Figure \ref{ruwrf}. Thus, 
we have $\boldsymbol{\Theta}=\psi(\mathcal{X}, \boldsymbol{\Phi})$, where $\boldsymbol{\Phi}$ denotes the parameters of the functional mapping, $\psi$. The redistribution kernel in (\ref{eq:spmk}) can now be rewritten as in (\ref{eq:spmkt}), yielding a multivariate nonstationary {kernel with functional parameters}.
\begin{equation}\label{eq:spmkt}
k^{(pq)}\left(\mathbf{s}, \mathbf{x} ; t, \boldsymbol{\Phi}\right)={\sigma_k^2}\exp \left(-\frac{\|\mathbf{s}-\mathbf{x}-\psi(\mathbf{X}_t, \boldsymbol{\Phi})^{{(p)}} t+\psi(\mathbf{X}_t, \boldsymbol{\Phi})^{{(q)}} (t+1)\|^2}{\ell_{\mathbbm{1}_{p=q}}^2}\right), 
\end{equation}
where $\psi(\mathbf{X}_t, \boldsymbol{\Phi})^{{(p)}}$ denotes the estimate of the advection vector at time $t$ and height $p$. 


{
By incorporating the 
kernel $k^{(pq)}(\mathbf{s}, \mathbf{x} ; t, \boldsymbol{\Phi})$ into the formulation of (\ref{eq:mide}), the resulting latent process model at location $\mathbf{s}$ and height $p$ can be expressed in its final form as in (\ref{eq:mide2}).
\begin{equation}\label{eq:mide2}
{
Y_t^{(p)}(\mathbf{s})=\sum_{\mathbf{x}=\mathbf{s}_1^*}^{\mathbf{s}_n^*} \sum_{q=1}^P \hat{k}^{(pq)}\left(\mathbf{s}, \mathbf{x} ; t, \boldsymbol{\Phi}\right) Y_{t-1}^{(q)}(\mathbf{x})+\eta_t^{(p)}(\mathbf{s}),}
\end{equation}
where $\hat{k}^{(pq)}\left(\mathbf{s}, \mathbf{x} ; t, \boldsymbol{\Phi}\right)$ is the kernel estimated by the data-driven mapping function $\psi(\cdot)$.}
{Note that many IDE-based models in the literature employ a reduced-rank basis expansion, which 
may exhibit limited performance in certain settings \citep{STEIN2014}. This issue is not as relevant in our context, since the goal is to describe the space-time dependence considering a fairly small number of observational locations. This enables us to work directly with the discretized IDE operator without relying on an explicit basis function expansion.}

\subsection{Deep learning as a physics extractor}\label{sec:dl} The remaining question is the choice of the mapping function $\psi(\cdot)$. Given the high-dimensional nature of $\mathcal{X}$, we propose a deep learning model to approximate the functional mapping $\psi(\cdot)$ and delineate the complex relationships between the exogenous image streams $\mathcal{X}=\{\mathbf{X}_t\}_{t=1}^T$ and the advection vectors $\boldsymbol{\Theta}$. In that regard, the embedded deep learning model acts as a ``physics extractor,'' by learning the advection dynamics of local wind fields from the wealth of exogenous information available to the forecaster. The proposed deep learning architecture in this work fuses AlexNet (a deep CNN-based model) \citep{krizhevsky2012imagenet} with a custom transformer architecture to fully process the spatial and temporal information in $\mathcal{X}$. Transformer models have shown tremendous promise in computer vision applications \citep{vaswani2023attention} and hence, are well-suited to process the stream of high-dimensional weather maps. 

\begin{figure}  
\centering  
\includegraphics[trim = 0cm .75cm 0cm 0cm, width = 1\linewidth]{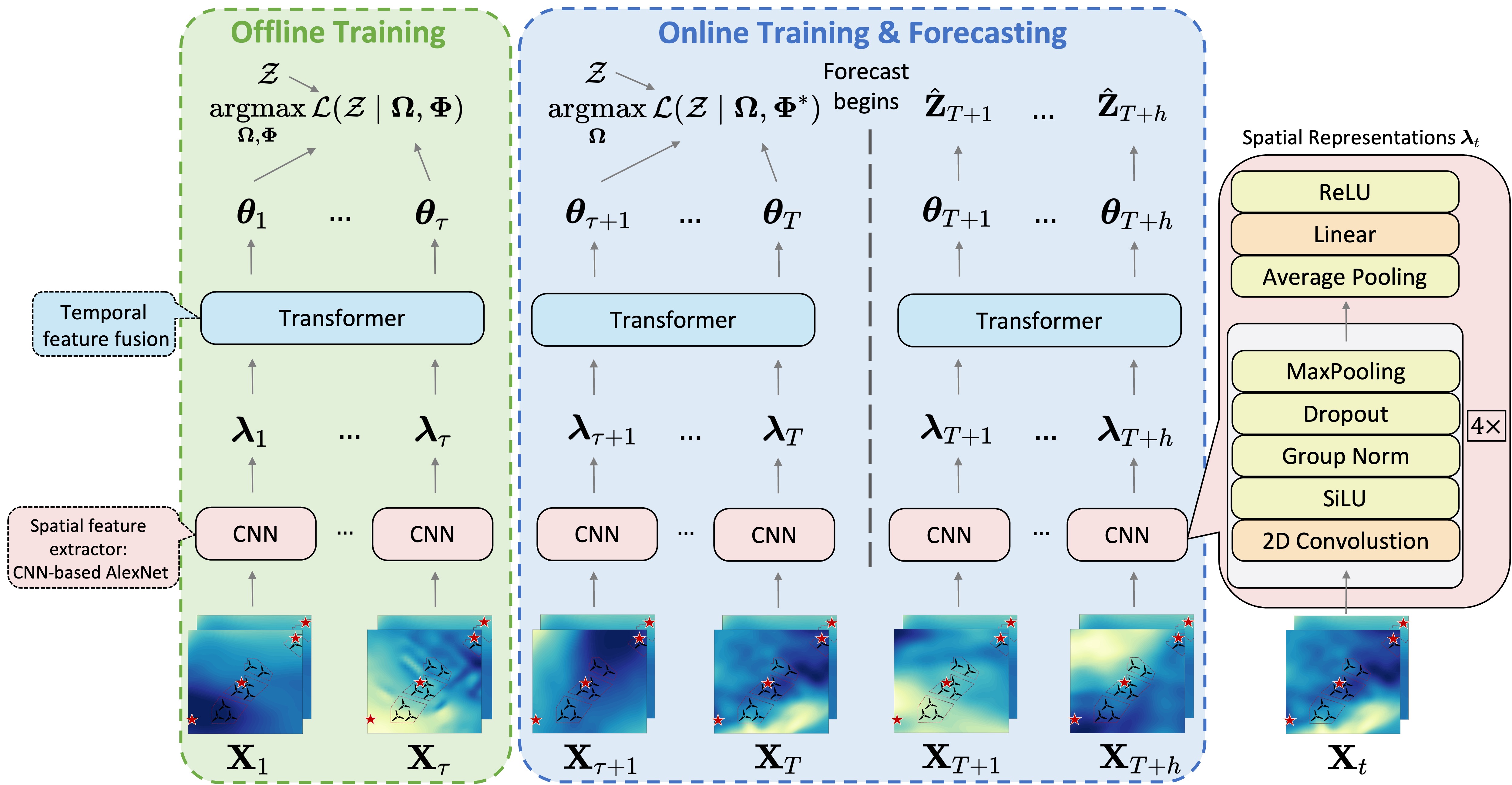}
\caption{Illustration of the DeepMIDE framework. During both offline and online training, the inputs to the network are the exogenous image streams $\mathbf{X}_t$ (e.g., Pressure, Temperature, etc) (offline $t \in 1, \hdots \tau$; online $t \in {\tau + 1, \hdots, T+h}$). 
In the offline phase, the network parameters $\boldsymbol{\Phi}$ are learned jointly with other statistical parameters $\boldsymbol{\Omega}$ by maximizing the likelihood $\mathcal{L}(\mathcal{Z}|\boldsymbol{\Omega}, \boldsymbol{\Phi})$. For the online phase, the deep learning model will be frozen, the network parameters $\boldsymbol{\Phi}$ are fixed and the statistical parameters $\boldsymbol{\Omega}$ will be continuously updated in light of new data. The structure of the deep CNN-based AlexNet is shown to the right. {The transformer model architecture, along with more details about the deep learning model, are deferred to the supplementary material document appended to this manuscript.}}
\label{fig-dl-model}  
\end{figure}

Figure \ref{fig-dl-model} provides a high-level illustration {of} DeepMIDE's workflow, comprising two phases: an offline training stage ($t \in \{1, \hdots, \tau\}$) and an online training and forecasting stage ($t \in \{\tau+1, \hdots, T+h\}$), {where $\tau$ is the number of timestamps in the offline training data, $T$ is the number of timestamps in the {offline and online} training data, and $h > 0$ is the forecast horizon.}
In the offline stage, the exogenous weather images $\{\mathbf{X}_t\}_{t=1}^\tau$ are processed through a sequence of deep CNNs to transform the high-dimensional image data into a condensed vector representation $\{\boldsymbol{\lambda}_t\}_{t=1}^{\tau}$. Subsequently, this sequence of vectors $\{\boldsymbol{\lambda}_t\}_{t=1}^\tau$, representing the compressed spatial features, is fed into a transformer model, which models the temporal and vertical dynamics across the spatial features, ultimately outputting the parameters $\boldsymbol{\Theta}$ that govern the advection dynamics for each corresponding time step. The CNN architecture used is shown to the right of Figure \ref{fig-dl-model}, whereas the transformer model architecture, along with more details about the deep learning model, are deferred to the supplementary material document appended to this manuscript. 

We denote by $\boldsymbol{\Phi}$ the entire set of deep learning parameters (of both the deep CNNs and the transformer), whereas the remaining statistical parameters of the IDE model are denoted by $\boldsymbol{\Omega}=\{\ell_{\mathbbm{1}_{p=q}}, \sigma_{\epsilon}$, $\sigma_{\eta}$, {$\sigma_k$,} $\ell_{\epsilon}$, $\ell_{\eta}\}$. Given $\mathcal{Z} = \{{\mathbf{Z}}_t\}$ $\forall t$, $\boldsymbol{\Phi}$, and $\boldsymbol{\Omega}$, the likelihood of DeepMIDE can be compactly expressed in closed-form as in (\ref{eq:likelihood}).
{
\begin{equation}
\mathcal{L}(\mathcal{Z}|\boldsymbol{\Omega}, \boldsymbol{\Phi}) = \mathcal{P}({\mathbf{Z}}_1|\boldsymbol{\Omega}, \boldsymbol{\Phi})\prod_{t}\mathcal{P}({\mathbf{Z}}_t|{\mathbf{Z}}_{1: t-1}, \boldsymbol{\Omega}, \boldsymbol{\Phi}),
\label{eq:likelihood}
\end{equation}
where ${\mathbf{Z}}_{1: t-1}\equiv \{{\mathbf{Z}}_1, \hdots, {\mathbf{Z}}_{t-1}\}$.} 
The exact form of $\mathcal{L}(\mathcal{Z}|\boldsymbol{\Omega}, \boldsymbol{\Phi})$ is discussed in Section \ref{sec:kf}. For the offline stage, we jointly optimize the full parameter set $\boldsymbol{\Phi}$ and $ \boldsymbol{\Omega}$ using stochastic gradient descent \citep{sgd}, with backpropagation used to compute the gradient $\nabla_{\boldsymbol{\Phi}, \boldsymbol{\Omega}} \mathcal{L}(\mathcal{Z}|\boldsymbol{\Phi}, \boldsymbol{\Omega})$, yielding an optimal set of parameters, denoted by $\boldsymbol{\Phi}^*$ and $\boldsymbol{\Omega}^*$. 
For the online stage, the deep learning model will be frozen, i.e., the network parameters $\boldsymbol{\Phi}$ are kept fixed at $\boldsymbol{\Phi}^*$ {and no longer updated, whereas the statistical parameters $\boldsymbol{\Omega}$ will be continuously estimated to maximize the likelihood in light of new observations}. To issue forecasts, the incoming weather maps are fed directly to the pre-trained deep learning model to predict the advection vector parameters $\{\boldsymbol{\theta}_t\}_{t = T+1}^{T+h}$, which are then plugged back into the {multi-output} IDE to output a set of multi-height space-time forecasts for $h$-steps ahead 
denoted by $\hat{\mathbf{Z}}_{T+1}, \hdots, \hat{\mathbf{Z}}_{T+h}$. An obvious advantage of the online/offline structure is that the deep learning model needs to be trained only once, 
thereby significantly reducing the computational burden in the online phase, where the model can quickly apply the pre-learnt network parameters to newly revealed weather maps.

\subsection{Inference and forecasting using DeepMIDE} \label{sec:kf}
As in single-output IDE models, we can leverage the Kalman Filter (KF) machinery to perform inference on the process and obtain the forecasting and filtering distributions. By writing (\ref{eq:mide2}) in matrix format, we have: 
\begin{equation}
\mathbf{Y}_t = \mathbf{K}_t\mathbf{Y}_{t-1}+\boldsymbol{\eta}_t, 
\label{eq:process}
\end{equation}
where $\mathbf{K}_t$ comprises all the pairwise evaluations of the redistribution kernel {in (\ref{eq:spmkt}). The dependence of $\mathbf{K}_t$ on $\mathbf{X}_t$ is dropped for notational convenience.}
From there, the forecast distribution conditioned on all past measurements 
is Gaussian, 
i.e., $\mathbf{Y}_t \mid\ {\mathbf{Z}}_{1: t-1} \sim \mathcal{N}\left(\mathbf{Y}_{t \mid t-1}, \mathbf{P}_{t \mid t-1}\right)$ such that $\mathbf{Y}_{t \mid t-1}$ and $\mathbf{P}_{t \mid t-1}$ are expressed as in (\ref{eq:forecast_dist}). 
\begin{equation}
\begin{aligned}
\mathbf{Y}_{t \mid t-1}&\equiv \mathbb{E}\left(\mathbf{Y}_t \mid {\mathbf{Z}}_{1: t-1}\right)=\mathbf{K}_t \mathbf{Y}_{t-1 \mid t-1}, \\
\mathbf{P}_{t \mid t-1}&\equiv \mathbb{E}[(\mathbf{Y}_{t}-\mathbf{Y}_{t \mid t-1})(\mathbf{Y}_{t}-\mathbf{Y}_{t \mid t-1})^{\top} \mid {\mathbf{Z}}_{1: t-1}] \\
& =\Sigma_{\boldsymbol{\eta}}+\mathbf{K}_t \mathbf{P}_{t-1 \mid t-1} \mathbf{K}_t^{\top} .
\end{aligned}
\label{eq:forecast_dist}
\end{equation}
The filtering distribution is also Gaussian $\mathbf{Y}_t \mid {\mathbf{Z}}_{1: t} \sim \mathcal{N}\left(\mathbf{Y}_{t \mid t}, \mathbf{P}_{t \mid t}\right)$ such that $\mathbf{Y}_{t \mid t}$ and $\mathbf{P}_{t \mid t}$ are expressed as in (\ref{eq:filtering_dist}). 
\begin{equation}
\begin{aligned}
\mathbf{Y}_{t \mid t}&\equiv \mathbf{Y}_{t \mid t-1}+\mathbf{A}_t\left({\mathbf{Z}}_t-\mathbf{H}_t \mathbf{Y}_{t \mid t-1}\right), \\
\mathbf{P}_{t \mid t}&\equiv\left(\mathbf{I}-\mathbf{A}_t \mathbf{H}_t\right) \mathbf{P}_{t \mid t-1}, \\
\mathbf{A}_t &\equiv \mathbf{P}_{t \mid t-1} \mathbf{H}_t^{\top}\left(\mathbf{H}_t^{\top} \mathbf{P}_{t \mid t-1} \mathbf{H}_t+\Sigma_{\boldsymbol{\epsilon}}\right)^{-1}. \\
\end{aligned}
\label{eq:filtering_dist}
\end{equation} 
From there, the explicit log-likelihood of DeepMIDE, previously presented in a compressed form in (\ref{eq:likelihood}), can now be written as in (\ref{eq:loglikelihood}).
\begin{equation}
\begin{aligned}
\log\mathcal{L}(\mathcal{Z}|\boldsymbol{\Omega}, \boldsymbol{\Phi}) = &
\sum_{t}\left(-\frac{mG}{2} \log (2 \pi) -\frac{1}{2} \log (|\mathbf{H}_t \mathbf{P}_{t|t-1} \mathbf{H}_t^\top +{\Sigma_{\boldsymbol{\epsilon}}}
|) \right. \\
& \left. -\frac{1}{2}\left({\mathbf{Z}}_t-\mathbf{H}_t \mathbf{Y}_{t \mid t-1} \right)^{\top}\left(\mathbf{H}_t \mathbf{P}_{t|t-1} \mathbf{H}_t^\top +{\Sigma_{\boldsymbol{\epsilon}}}\right)^{-1}\left({\mathbf{Z}}_t-\mathbf{H}_t \mathbf{Y}_{t \mid t-1}\right) \right),
\label{eq:loglikelihood}
\end{aligned}
\end{equation}
where $|\cdot|$ is the matrix determinant. The distribution of the $h$-step latent process forecasts $\hat{\mathbf{Y}}_{T+h} \mid {\mathbf{Z}}_{1: T} \sim \mathcal{N}(\hat{\mathbf{Y}}_{T+h \mid T+h-1}, \hat{\mathbf{P}}_{T+h \mid T+h-1})$ can be expressed as: 
\begin{equation}
\begin{aligned}
\hat{\mathbf{Y}}_{T+h \mid T+h-1}&=\mathbf{K}_{T+h} \hat{\mathbf{Y}}_{T+h-1 \mid T+h-2}, \\
\hat{\mathbf{P}}_{T+h \mid T+h-1}&=\Sigma_{\boldsymbol{\eta}}+\mathbf{K}_{T+h} \hat{\mathbf{P}}_{T+h-1 \mid T+h-2} \mathbf{K}_{T+h}^{\top}.
\end{aligned}
\end{equation}
The final distribution of the $h$-step forecasts $\hat{\mathbf{Z}}_{T+h} \mid {\mathbf{Z}}_{1: T} \sim \mathcal{N}(\hat{\mathbf{Z}}_{T+h \mid T+h-1}, \hat{\mathbf{C}}_{T+h \mid T+h-1})$ can be expressed as in (\ref{eq:findist}). 
\begin{equation}\label{eq:findist}
\begin{aligned}
\hat{\mathbf{Z}}_{T+h \mid T+h-1}&= \mathbf{H}_{T+h}\hat{\mathbf{Y}}_{T+h \mid T+h-1}, \\
\hat{\mathbf{C}}_{T+h \mid T+h-1}&=\Sigma_{\boldsymbol{\epsilon}}+\mathbf{H}_{T+h} \hat{\mathbf{P}}_{T+h \mid T+h-1} \mathbf{H}_{T+h}^{\top},
\end{aligned}
\end{equation}
{
where $\hat{\mathbf{Y}}_{T+h \mid T+h-1}$ and $\hat{\mathbf{P}}_{T+h \mid T+h-1}$ denote the forecast vector of the latent state process and its covariance matrix, whereas 
$\hat{\mathbf{Z}}_{T+h \mid T+h-1}$ and $\hat{\mathbf{C}}_{T+h \mid T+h-1}$ denote the forecast vector of the observed measurements and its covariance matrix. }

\section{Real-world experiments and discussions}\label{results}
{In this work, we focus on $2$-hour-ahead forecasts at a $10$-minute resolution. Accurate forecasts at this look-ahead horizon are used in practice to inform important wind farm and power system operations, including resource commitment, dispatch, and market participation \citep{nyisortc}}. 
We train and evaluate DeepMIDE at the three locations where data is available: E05N, E06, and ASOW6. The data spans a total duration of approximately $8.5$ months, so we use six months for offline training, and the remainder is used for online training and forecast evaluation, carried out in a rolling fashion. 
{Careful handling of the time index in the redistribution kernel in (\ref{eq:spmkt}) is warranted, considering that the time index can, in theory, grow without bound leading to inconsistent physical behavior. Hence, training and forecasting are performed in rolling windows with normalized time, where $t$ is reset at the start of each window. Similarly, the advection parameters are bounded to ensure physically meaningful values.}
{For each roll, 
we make $2$-hour forecasts, slide by six hours, and repeat the whole process. 
This corresponds to $\mathcal{R} = 284$ rolls, yielding a total of $6$ forecasts/hour $\times$ $2$-hour horizon $\times$ $284$ rolls $\times$ $3$ spatial sites $\times$ $3$ heights = $30{,}672$ testing instances. In this paper, the observation mapping matrix $\mathbf{H}_t$ is assumed to be an identity matrix, i.e.,  $m = n = 3, G=P=3$, {such that the observation and process locations are the same, that is, $\mathbf{s}_i = \mathbf{s}^*_i$ $\forall i = 1, ..., n$. This is because our goal is to obtain forecasts at the same locations where the wind turbines are installed.}} We find that one week (7 days) of historical data is a sufficient online training data size to balance model fitting and computational efficiency.
The offline training takes approximetely $7$ hours using an \texttt{RTX4090} GPU with $46$ GB RAM, while the online training phase takes approximately 3 minutes.

We compare DeepMIDE against {six} benchmarks that are representative of the various forecasting approaches in the wind forecasting literature and practice.
\begin{itemize}
\item \textbf{PER}: The persistence model is widely regarded as a standard forecasting benchmark. PER assumes that the observed wind speed at the present time will persist into the future. It is implemented separately for each height and location. {Despite its simplicity, the performance of PER is often highly competitive at short-term horizons.} 

\item \textbf{ARIMA}: The Autoregressive Integrated Moving Average (ARIMA) model is a classical time series approach that considers temporal correlations, but overlooks spatial and vertical dependence. It is implemented for each location and height. All model parameters and coefficients are optimized using the \texttt{pmdarima} package in {Python}. 

\item \textbf{STGP}: The spatio-temporal Gaussian process model is a {prevalent} geostatistical approach {which considers correlations over space and time, but not height. For its covariance function, we use a convex combination of Schlater's Lagrangian kernel \citep{SchlatherCov} and a separable squared exponential kernel, as expressed in (\ref{eq:schlather}). 
\begin{equation} \label{eq:schlather}
\begin{split}
k^{\mathrm{STGP}}\bigl(\mathbf{s}, \mathbf{x}; t_1, t_2, \boldsymbol{\nu}, \boldsymbol{\Sigma}\bigr)
  ={}& \sigma_k^2\Biggl(
         \rho \,\frac{1}{\sqrt{\,
             \bigl|\mathbf{I}_{2\times 2} + 2\,\boldsymbol{\Sigma}(t_1-t_2)^2\bigr|
           \,}}\,
         \exp\Bigl\{ 
           -(\mathbf{s} -\mathbf{x} - \boldsymbol{\nu}(t_1-t_2))^{T}\,
           \\&\bigl(\mathbf{I}_{2\times 2} + 2\,\boldsymbol{\Sigma}(t_1-t_2)^2\bigr)^{-1}\,
           (\mathbf{s}-\mathbf{x} - \boldsymbol{\nu}(t_1-t_2))
         \Bigr\} +\\&
        (1 - \rho)\,
         \Bigl(
           \exp\bigl(-\tfrac{\|\mathbf{s}-\mathbf{x}\|^2}{\ell_{\mathbf{s}}^2}\bigr)\,
           \exp\bigl(-\tfrac{\| t_1 - t_2\|^2}{\ell_{{t}}^2}\bigr)
         \Bigr)
       \Biggr),
\end{split}
\end{equation}
where $\rho$ is an asymmetry parameter denoting the strength of asymmetry, $\ell_{\mathbf{s}}^2$ and $\ell_{t}^2$ are spatial and temporal length-scale parameters, respectively. The advection vectors in Schlather's Lagrangian kernel are assumed to be random vectors following a bivariate Gaussian distribution for which the mean $\boldsymbol{\nu}$ and covariance $\boldsymbol{\Sigma}$ are estimated using the wind velocity forecasts from the NWP model. A separate STGP model is trained for each height using the \texttt{fitrgp} package in 
{MATLAB}. Hence, this space-time model, while capable of capturing complex space-time correlation structures, does not account for vertical dependencies and interactions.} 

\item \textbf{DeepAR}: {The Deep Autoregressive model (DeepAR) is the space-time model proposed by \cite{deepar} in which a sequence of RNNs is linked in an autoregressive-based manner. 
Specifically, the historical observations are fed into a deep recurrent neural network with {Long Short-Term Memory} (LSTM) cells, which directly outputs the parameters of a statistical likelihood function (e.g., Gaussian). Although it is fundamentally deep-learning-based (the main workhorse of making forecasts is an RNN), the relevance of this benchmark is that\textemdash similar to DeepMIDE\textemdash it falls under the general category of statistical deep learning methods.} We train the DeepAR using the \texttt{GluonTS} package in Python, and follow the authors' suggestions for setting the parameters \citep{deepar}. 
{\item \textbf{DeepIDE-H}: DeepIDE-H is an adaptation of the single-output DeepIDE model from \cite{zammit2020deep}, where height is incorporated as an additional spatial coordinate. In that regard, DeepIDE-H combines the horizontal and vertical dimensions into a three-dimensional spatial domain. The redistribution kernel for DeepIDE-H is denoted as $k^{\text{IDE-H}}$ and is defined in (\ref{eq:ide-h}) as the product of an asymmetric kernel for the horizontal spatial components and a squared exponential kernel for the vertical component. 
{Here, $\boldsymbol{\theta}_t$ and $\gamma_t$ are estimated using a deep learning model that shares the same architecture as the one used in DeepMIDE. However,} unlike DeepMIDE, this benchmark is a single-output model as it does not distinguish height as a physically meaningful vertical axis, {and treats it as a spatial coordinate.} 
{
\begin{equation} \label{eq:ide-h}
k^{\text{IDE-H}}\left(\mathbf{s}, \mathbf{x} ; \kappa_e, \kappa_f, \boldsymbol{\theta}_t, \gamma_t\right)=\sigma_k^2\exp \left(-\frac{\left\|\mathbf{s}-\mathbf{x}-\boldsymbol{\theta}_t\right\|^2}{\ell_1^2}\right) \exp \left(-\frac{\|\kappa_e-\kappa_f-\gamma_t\|^2}{\ell_2^2}\right),
\end{equation}
}
{where $\kappa_e$ and $\kappa_f$ are the actual altitude values for height $e$ and $f$, respectively, whereas $\ell_1^2$ and $\ell_2^2$ are length-scale parameters.}}
{\item \textbf{NWP}: This is the offshore wind speed forecast from the physics-based numerical weather prediction (NWP) model: RU-WRF, which is introduced in Section \ref{ru-wrf}. The forecasts are available in hourly resolution for each height and location. 
}
\end{itemize}

{The above six benchmarks can be grouped into the following categories representing different paradigms for wind forecasting: (\textit{i}) Statistical methods: PER, ARIMA, and STGP; (\textit{ii}) Statistical deep learning methods: DeepAR and DeepIDE-H; and (\textit{iii}) Physics-based methods, comprising the base forecasts from the NWP model.} We present the forecasting experiments in Section \ref{results_forecast}, followed by some in-depth discussions about the performance of DeepMIDE. 

\subsection{Forecasting results} \label{results_forecast}
\begin{figure}  
\centering  
\includegraphics[trim = 0cm .75cm 0cm 0cm, width = 1\linewidth]{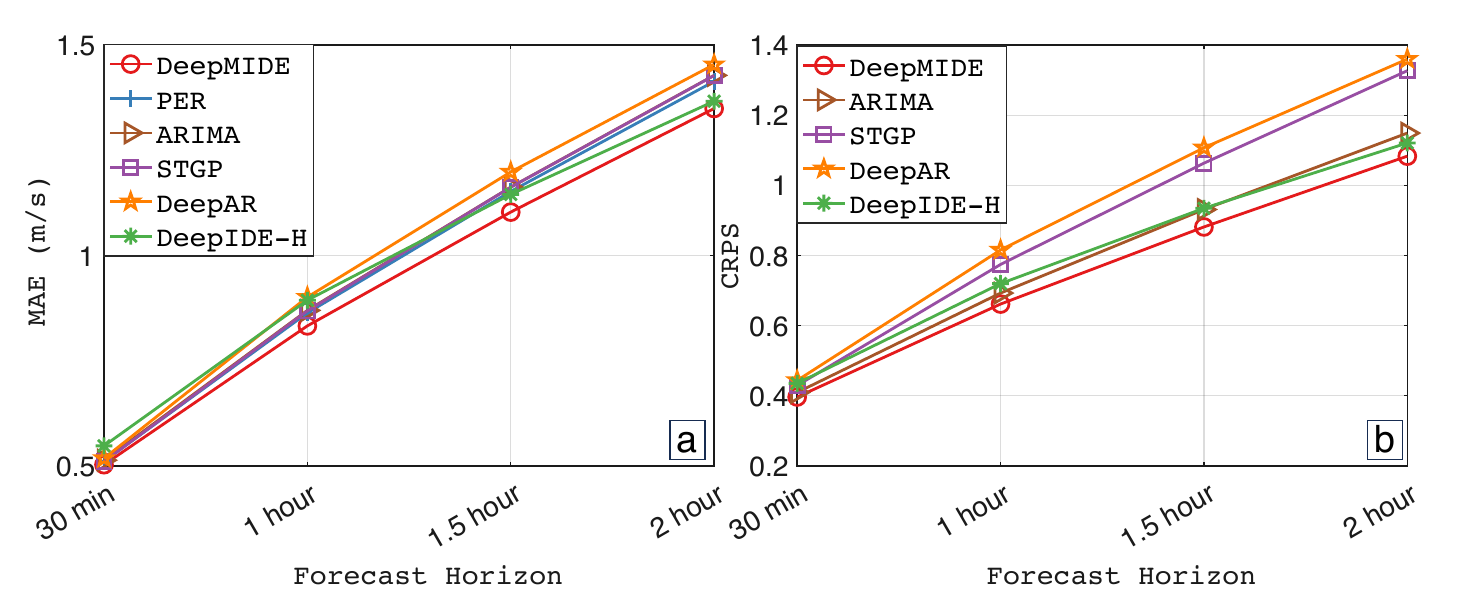}
\caption{{(a) MAE versus the forecast horizon for all models. (b) CRPS versus the forecast horizon for all models (excluding PER due to poor performance for probabilistic forecasts).}}
\label{fig: mae1}  
\end{figure}

Table \ref{tab:mae1} shows the spatially averaged mean absolute error (MAE) values for all models at different forecast horizons for the three heights considered, namely: $100$m, $140$m, and $180$m. Specifically, the MAE for a representative {method} $\mathcal{B}$ {(be it PER, ARIMA, STGP, DeepAR, NWP, DeepIDE-H, or DeepMIDE) is} denoted by $\text{MAE}^{(g)}_h(\mathcal{B})$, {and} is calculated as in (\ref{eq:mae}).
\begin{equation}\label{eq:mae}
\text{MAE}^{(g)}_h(\mathcal{B}) = \frac{1}{\mathcal{R} \times m}\sum_{i=1}^{\mathcal{R}}\sum_{j=1}^m (|{Z^{(g)}_{T+h, i}(\mathbf{s}_j)} - {\hat{Z}^{(g)}_{T+h, i}(\mathbf{s}_j)}|),
\end{equation}
where ${Z^{(g)}_{T+h, i}(\mathbf{s}_j)}$ and $\hat{Z}^{(g)}_{T+h, i}(\mathbf{s}_j)$ denote the actual wind speed and forecasts at location $\mathbf{s}_j$ and height $g$ for forecasting horizon $h$ and $i$th forecasting roll, respectively.
The percentage reduction {in forecast error} is computed using the average of all absolute errors taken across all forecast rolls and locations for the two methods being compared, as expressed in (\ref{eq:imp}).
\begin{equation}\label{eq:imp}
\text{IMP}(\mathcal{B}^*, \mathcal{B}) = 100 \times (1-\frac{\text{MAE}^{(g)}_h(\mathcal{B}^*)}{\text{MAE}^{(g)}_h(\mathcal{B})}),
\end{equation}
where $\mathcal{B}^*$ is the proposed DeepMIDE model, while $\mathcal{B}$ is any benchmark {model}. 
Figure \ref{fig: mae1}(a) shows the MAE across all locations and heights ({We exclude NWP due to its poor predictive performance}). 
A first glance at Table \ref{tab:mae1} and Figure \ref{fig: mae1}(a) shows that DeepMIDE consistently outperforms all benchmarks across all horizons and heights {(100m, 140m, and 180m), with improvements ranging from approximately $3.6$\% to $4.8$\% compared to other statistical models (PER, ARIMA, STGP), and about {$3.9$\%} to $7.2$\% compared to statistical deep learning models (DeepAR and DeepIDE-H). The physics-based NWP forecasts from RU-WRF exhibit significantly higher MAE. This is unsurprising considering that NWP models are not competitive at these short look-ahead horizons \citep{ye2023ieee}. 
Despite that, NWP data remains valuable in conveying physically meaningful information about advection dynamics, which are harnessed by DeepMIDE's embedded deep learning architecture.}

{It is also worth noting that, 
at the $30$-minute horizon ($0.5$ hours ahead), where it is generally hard to beat simpler statistical models such as persistence and autoregressive methods, DeepMIDE maintains a small but meaningful margin of improvement, indicating it can capture immediate wind fluctuations, by virtue of its dynamical statistical representation. As the lead time increases, all models experience some degradation in accuracy, but DeepMIDE’s error grows more slowly than other benchmarks. DeepIDE-H seems to overtake PER as the second-best performing model as the forecast horizon extends ($h > 1$ hour ahead). Yet, it is still outperformed by DeepMIDE as clearly shown in Figure \ref{fig: mae1}(a), further confirming the merit of the multi-output approach pursued herein. At the $2$-hour ahead horizon, DeepMIDE still maintains a noticeably lower MAE than any other method, while the error of the DeepAR, for example, increases more rapidly with forecast horizon. DeepAR, although also a statistical deep learning model, shows lower performance. Unlike the proposed DeepMIDE, which is fundamentally a statistical model augmented by deep learning as a ``physics extractor'', DeepAR is fundamentally a deep learning-based model and may require more training data to get better performance. Trailing behind is STGP, which does not appear to be capable of matching the performance of simpler statistical models in very short-term horizons (namely, PER and ARIMA), nor the more sophisticated Deep IDE models at longer horizons.} {Yet, STGP still outperforms the DeepAR and NWP models, likely due to its expressive kernel structure shown in (\ref{eq:schlather}).}

{Another interesting observation from Table \ref{tab:mae1} is that DeepMIDE achieves its greatest improvements at the 140m height compared to other benchmarks 
suggesting an advantage in leveraging central heights by effectively capturing advection and vertical cross-dependencies.} We also find that, on average, all models (including DeepMIDE) have slightly higher MAE values at higher altitudes. However, this is likely an artifact of the wind speeds being stronger as the altitude increases.

\begin{table}[htpb]\caption{{Average MAE values (aggregated in 30-min intervals) versus the forecast horizon $h$ for all models, at different heights, namely: $100$m, $140$m, and $180$m. Bold-faced and underlined values denote the best and second-best performing models, respectively.}}\label{tab:mae1}
\resizebox{\columnwidth}{!}{%
\begin{tabular}{|cc|ccccccc|}
\hline
\multicolumn{2}{c}{}    &   \textit{Physics-Based}   &   \multicolumn{3}{c}{\textit{Statistical Methods}} &   \multicolumn{3}{c}{\textit{Statistical Deep Learning Methods}}  \\ \hline
Location & $h$ (hour) & NWP    & PER   & ARIMA & STGP  & DeepAR & DeepIDE-H & DeepMIDE \\ \hline
         & 0.5        & 1.644  & 0.504 & \underline{0.503} & 0.504 & 0.510  & {0.541}     & \textbf{0.497} \\
         & 1          & 1.459  & \underline{0.844} & 0.848 & 0.848 & 0.882  & {0.874}     & \textbf{0.816} \\
100$m$   & 1.5        & 1.453  & 1.092 & 1.102 & 1.103 & 1.144  & {\underline{1.084}}     & \textbf{1.046} \\
         & 2          & 1.412  & 1.338 & 1.351 & 1.354 & 1.389  & {\underline{1.289}}     & \textbf{1.280} \\
         \cdashline{1-9}
         & Avg        & 1.492  & \underline{0.944} & 0.951 & 0.952 & 0.981  & {0.947}     & \textbf{0.910} \\
         & IMP($\%$)  & 39.008 & 3.602 & 4.311 & 4.412 & 7.238  & {3.907}     &         \\ \hline
         
         & 0.5        & 1.751  & \underline{0.507} & 0.511 & \underline{0.507} & 0.515  & {0.543}     & \textbf{0.499} \\
         & 1          & 1.556  & \underline{0.860} & 0.865 & 0.866 & 0.889  & {0.890}     & \textbf{0.826} \\
140$m$   & 1.5        & 1.548  & \underline{1.155} & 1.164 & 1.166 & 1.199  & {1.149}     & \textbf{1.105} \\
         & 2          & 1.517  & 1.421 & 1.436 & 1.435 & 1.462  & {\underline{1.373}}     & \textbf{1.354} \\ \cdashline{1-9}
         & Avg        & 1.593  & \underline{0.986} & 0.994 & 0.993 & 1.016  & {0.989}     & \textbf{0.946} \\
         & IMP($\%$)  & 40.615 & 4.057 & 4.829 & 4.733 & 6.890  & {4.348}     &         \\ \hline
         
         & 0.5        & 1.822  & 0.521 & 0.526 & \underline{0.520} & 0.528  & {0.558}     & \textbf{0.513} \\
         & 1          & 1.608  & \underline{0.886} & 0.895 & 0.891 & 0.932  & {0.917}     & \textbf{0.856} \\
180$m$   & 1.5        & 1.612  & {1.210} & 1.222 & 1.218 & 1.253  & {\underline{1.203}}     & \textbf{1.158} \\
         & 2          & 1.579  & 1.482 & 1.497 & 1.495 & 1.511  & {\underline{1.438}}     & \textbf{1.414} \\
         \cdashline{1-9}
         & Avg        & 1.655  & \underline{1.025} & 1.035 & 1.031 & 1.056  & {1.029}     & \textbf{0.985} \\
         & IMP($\%$)  & 40.483 & 3.902 & 4.831 & 4.462 & 6.724  & {4.276}     &         \\ \hline
\end{tabular}%
}
\end{table}

\begin{table}[]\caption{{Average MAE values (aggregated in 30-min intervals) versus the forecast horizon $h$ for all models, at different locations, namely: E05N, E06, and ASOW6. Bold-faced and underlined values denote the best and second-best performing models, respectively.}}\label{tab:mae2}
\resizebox{\columnwidth}{!}{%
\begin{tabular}{|cc|ccccccc|}
\hline
\multicolumn{2}{c}{}    &   \textit{Physics-Based}   &   \multicolumn{3}{c}{\textit{Statistical Methods}} &   \multicolumn{3}{c}{\textit{Statistical Deep Learning Methods}}  \\ \hline
Location & $h$ (hour) & NWP    & PER   & ARIMA & STGP  & DeepAR & DeepIDE-H & DeepMIDE \\ \hline
         & 0.5        & 1.657  & 0.477 & \underline{0.476} & 0.475 & 0.478  & {0.493}     & \textbf{0.473} \\
         & 1          & 1.491  & 0.811 & \underline{0.803} & 0.815 & 0.831  & {0.808}     & \textbf{0.797} \\
E05N     & 1.5        & 1.526  & 1.103 & 1.092 & 1.109 & 1.123  & {\textbf{1.052}} & \underline{1.083} \\
         & 2          & 1.530  & 1.360 & 1.358 & 1.371 & 1.383  & {\textbf{1.282}} & \underline{1.341} \\
         \cdashline{1-9}
         & Avg        & 1.551  & 0.938 & 0.932 & 0.942 & 0.954  & {\textbf{0.909}} & \underline{0.923} \\
         & IMP($\%$)  & 40.490 & 1.599 & 0.966 & 2.017 & 3.250  & {-1.540}     &         \\ \hline
         
         & 0.5        & 1.766  & 0.513 & 0.524 & 0.514 & \underline{0.510}  & {0.534}     & \textbf{0.489} \\
         & 1          & 1.533  & 0.933 & 0.956 & 0.939 & 0.946  & {\underline{0.880}}     & \textbf{0.856} \\
E06      & 1.5        & 1.498  & 1.219 & 1.245 & 1.234 & 1.235  & {\textbf{1.055}} & \underline{1.089} \\
         & 2          & 1.459  & 1.485 & 1.513 & 1.506 & 1.496  & {\textbf{1.247}} & \underline{1.315} \\
         \cdashline{1-9}
         & Avg        & 1.564  & 1.037 & 1.059 & 1.048 & 1.047  & {\textbf{0.929}} & \underline{0.937} \\
         & IMP($\%$)  & 40.090 & 9.643 & 11.520 & 10.592 & 10.506 & {-0.861}     &         \\ \hline
         
         & 0.5        & 1.794  & 0.542 & \textbf{0.540} & \underline{0.541} & 0.565  & {0.616}     & 0.547 \\
         & 1          & 1.598  & \underline{0.847} & 0.850 & 0.851 & 0.927  & {0.993}     & \textbf{0.844} \\
ASOW6    & 1.5        & 1.589  & \underline{1.134} & 1.150 & 1.144 & 1.238  & {1.328}     & \textbf{1.138} \\
         & 2          & 1.518  & \underline{1.397} & 1.413 & 1.406 & 1.483  & {1.572}     & \textbf{1.393} \\
         \cdashline{1-9}
         & Avg        & 1.625  & \textbf{0.980} & 0.988 & \underline{0.986} & 1.053  & {1.127}     & \textbf{0.980} \\
         & IMP($\%$)  & 39.692 & 0.000 & 0.810 & 0.609 & 6.933  & {13.044}    &         \\ \hline
\end{tabular}%
}
\end{table}

Table \ref{tab:mae2} shows the MAE values, {averaged over height}, 
across the three spatial sites considered: E05N, E06, and ASOW6, which is denoted as $\text{MAE}^{(\mathbf{s})}_h(\mathcal{B})$, and calculated as in (\ref{eq:mae2}). 
\begin{equation} \label{eq:mae2}
\text{MAE}^{(\mathbf{s})}_h(\mathcal{B}) = \frac{1}{\mathcal{R} \times G}\sum_{i=1}^{\mathcal{R}}\sum_{g=1}^G (|{Z^{(g)}_{T+h, i}(\mathbf{s})} - {\hat{Z}^{(g)}_{T+h, i}(\mathbf{s})}|).
\end{equation}
{
Interestingly, this time, DeepIDE-H achieves competitive accuracy compared to DeepMIDE at the central site E06 and downstream site E05N, especially at longer horizons ($h > 1$ hour ahead). This suggests that for those locations, the horizontal neighbours provide sufficient spatial context for both models' horizontal advection kernel, wherein treating height as an independent spatial coordinate (as DeepIDE-H does) is nearly sufficient to capture the local wind field advection. However, this observation does not generalize to other sites. At the upstream boundary location ASOW6, DeepIDE-H performs poorly relative to DeepMIDE.
We postulate that horizontal information is scarce, and the forecast depends on propagation between heights, where DeepIDE-H uses a separable vertical kernel, so it cannot redirect information up or down when shear changes.
This suggests that the kernel structure in DeepIDE-H, which treats height as an additional spatial coordinate, can capture horizontal advection patterns to some extent, but might not fully capture the complex cross-advection dynamics between different heights.}

{In terms of all other benchmarks, the location-specific analysis in Table \ref{tab:mae2} further confirms DeepMIDE's superiority, showing the lowest average overall MAE across all locations. 
The maximal improvement appears at E06 (in the range of $\sim$\hspace{-0.02mm}$10$-$12$\% reduction in MAE). We postulate that this may be due to E06 being at a central location, in between the two other spatial sites, making the gain from including neighborhood information greater. 
}

{
Evaluating probabilistic forecasting performance, Figure \ref{fig: mae1}(b) presents the Continuous Ranked Probability Score (CRPS) for all models across the forecast horizon (We exclude PER due to poor performance for probabilistic forecasts). DeepMIDE consistently achieves the lowest CRPS values, indicating superior probabilistic forecast accuracy compared to all benchmarks. In terms of model adequacy quantified by the $R^2$ coefficient, DeepMIDE achieves the highest score ($R^2=0.867$), followed by DeepIDE-H ($R^2=0.864$), PER ($R^2=0.851$), STGP ($R^2=0.848$), ARIMA ($R^2=0.848$), DeepAR ($R^2=0.844$), and NWP ($R^2=0.687$). 
These results further highlight the predictive capabilities of DeepMIDE in both deterministic and probabilistic forecasting.
}

\begin{figure}[htpb] 
\centering   
\includegraphics[trim = .5cm .75cm 0.5cm 0cm, width = 1\linewidth]{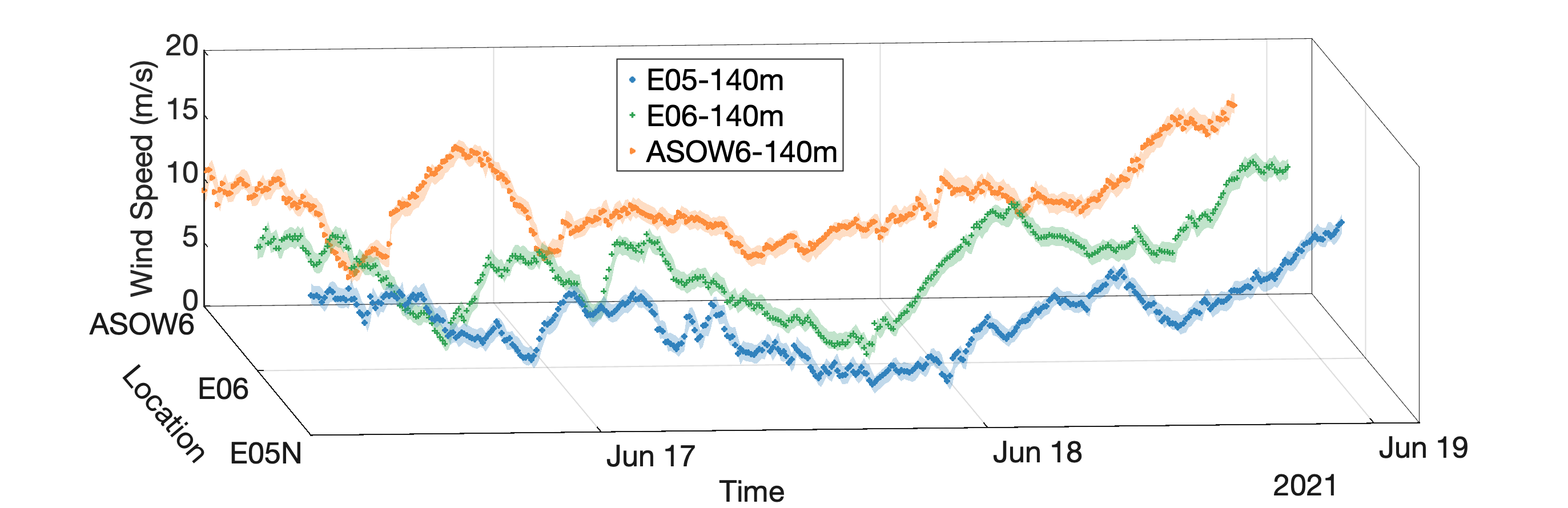}
\caption{Time series of $1$-step ahead forecasts from DeepMIDE on top of the actual wind speed observations at the hub height of $140$m for all three locations ($95$\% forecast intervals shown).}
\label{fig: ts_fore}  
\end{figure}

Figure \ref{fig: ts_fore} shows the time series of the one-step wind speed forecasts from DeepMIDE on top of the actual measurements at the hub height of $140$m, during a select time interval in June 2021. The time series of the forecasts show faithful agreement with the actual measurements for all three locations. The corresponding $95$\% forecast intervals visually show satisfactory coverage and sharpness. This probabilistic nature of IDE-based models make them well-suited to inform subsequent operational decision-making under uncertainty for offshore wind farms \citep{Pinson2013, papadopoulos2024stochos}. 

\begin{figure}  
\centering  
\includegraphics[trim = 0cm .75cm 0cm 0cm, width = 1\linewidth]{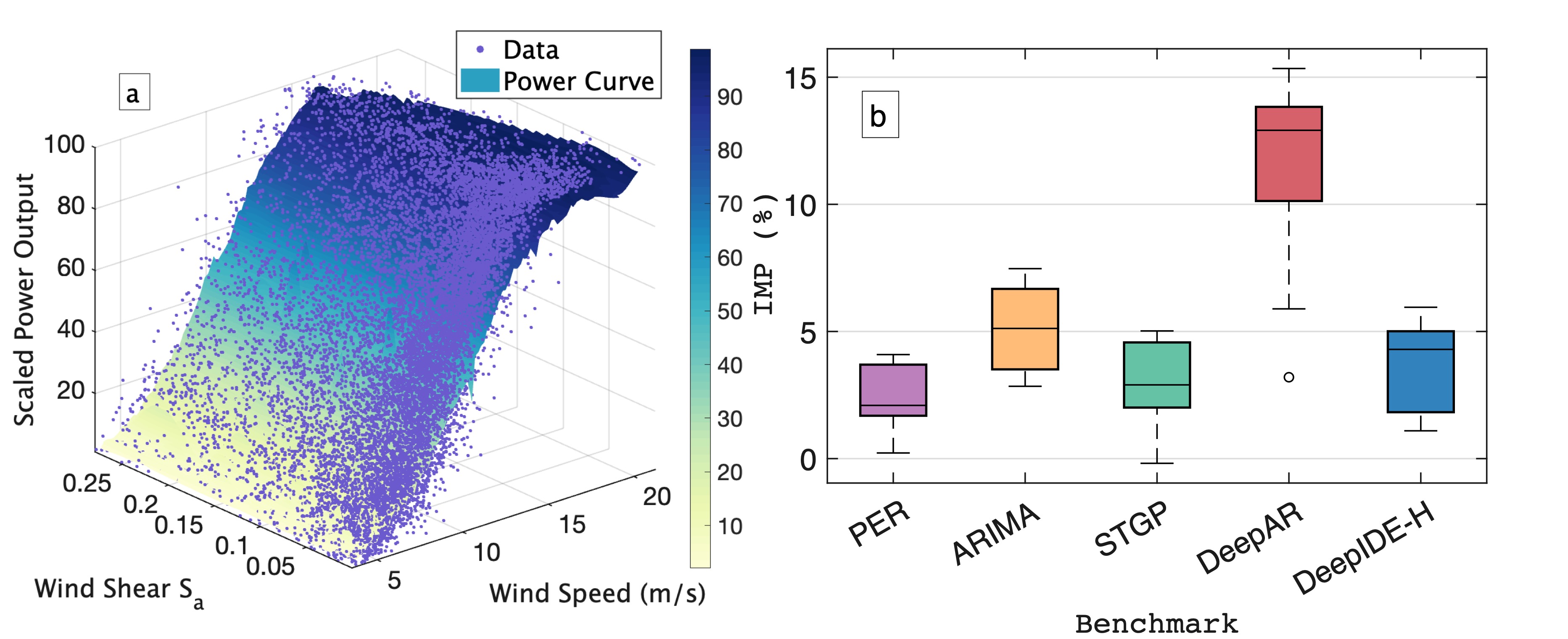}
\caption{(a) Fitted wind power curve with hub height wind speed and above-hub wind shear $S_a$, (b) the boxplots of improvement for all forecasting rolls, horizons, locations, and heights in wind power.}
\label{fig: pc_wp}  
\end{figure}

Wind speed is the major determinant of wind power, and hence, we would like to illustrate the value of our approach for improving wind power prediction accuracy. To do so, we need to convert the wind speed forecasts into the wind power domain. As there are no operational wind farms in the NY/NJ Bight region yet, we leverage an external set of operational data from a wind farm in the United States to construct a statistical wind power curve which will be used for wind-to-power conversion \citep{ding2019data}. The power curve is multivariate and it takes two inputs: the hub-height wind speed and the vertical wind shear. The latter reflects the vertical change in wind speed. Let the wind shear be denoted by $S$, then an estimate for wind shear is computed as in (\ref{eq:shear}). 
\begin{equation}
    S = \frac{\log{(Z_2/Z_1)}}{\log{{(b_1/b_2)}}},
    \label{eq:shear}
\end{equation}
where $Z_2$ and $Z_1$ are the wind speeds measured at heights {$b_1$ and $b_2$}, respectively \citep{lee2015power}. For our data, wind speeds are measured at heights of $100$m, $140$m, and $180$m, with $140$m considered as the hub-height. Consequently, two wind shear values are calculated: an above-hub wind shear ($S_a$) using the $180$m/$140$m pair, and a below-hub wind shear ($S_b$) using the $140$m/$100$m pair. Figure \ref{fig: pc_wp}(a) visualizes the fitted wind power curve (using a multivariate GP model akin to the one proposed by \cite{golparvar2021surrogate}) with hub height wind speed and above-hub wind shear $S_a$. The power output is scaled to the $[0, 100]$ interval, where $100$ represents the maximum rated capacity. By applying the constructed power curve, we transform the wind speed observations, along with their forecasts from the {six} competing models into corresponding wind power values.

Figure \ref{fig: pc_wp}(b) presents the boxplots of the improvement across all forecasting rolls, horizons, locations, and heights in wind power. DeepMIDE significantly outperforms all its competitors, demonstrating an average improvement of {$5.47$\%} in wind power forecasting compared to other benchmarks. Enhanced accuracy in wind speed forecasts directly contributes to greater improvements in wind power predictions. Theoretically, wind speed is related to wind power through a cubic relationship, and hence, large improvements in forecasting wind speed are expected to yield even greater improvements in predicting wind power. 

\subsection{Discussion}
\begin{figure}[h]  
\centering  
\includegraphics[trim = 0.5cm .75cm 0.5cm 0cm, width = 1\linewidth]{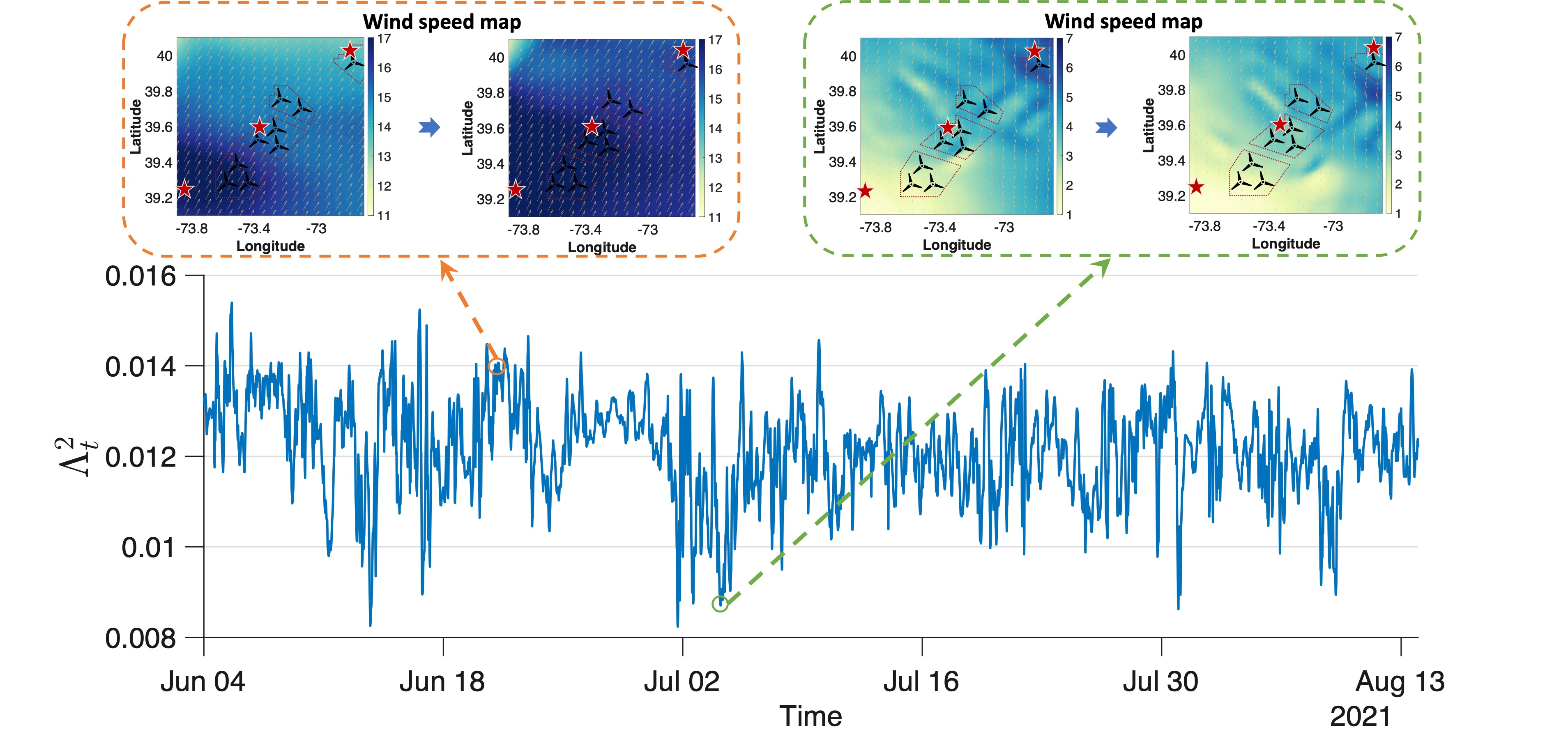}
\caption{Time series plot showing the norm of the predicted advection parameters at height $140$m, alongside spatial wind speed images from the numerical model at two consecutive {hourly} time steps, for two different time intervals: 21-Jun-2021 02:00:00 to 03:00:00 (orange rectangle, left) and from 04-Jul-2021 05:00:00 to 06:00:00 (green rectangle, right).}
\label{fig: adv_map}  
\end{figure}

\begin{figure}[h]  
\centering  
\includegraphics[trim = 0cm .25cm 0cm 0.25cm, width = 0.7\linewidth]{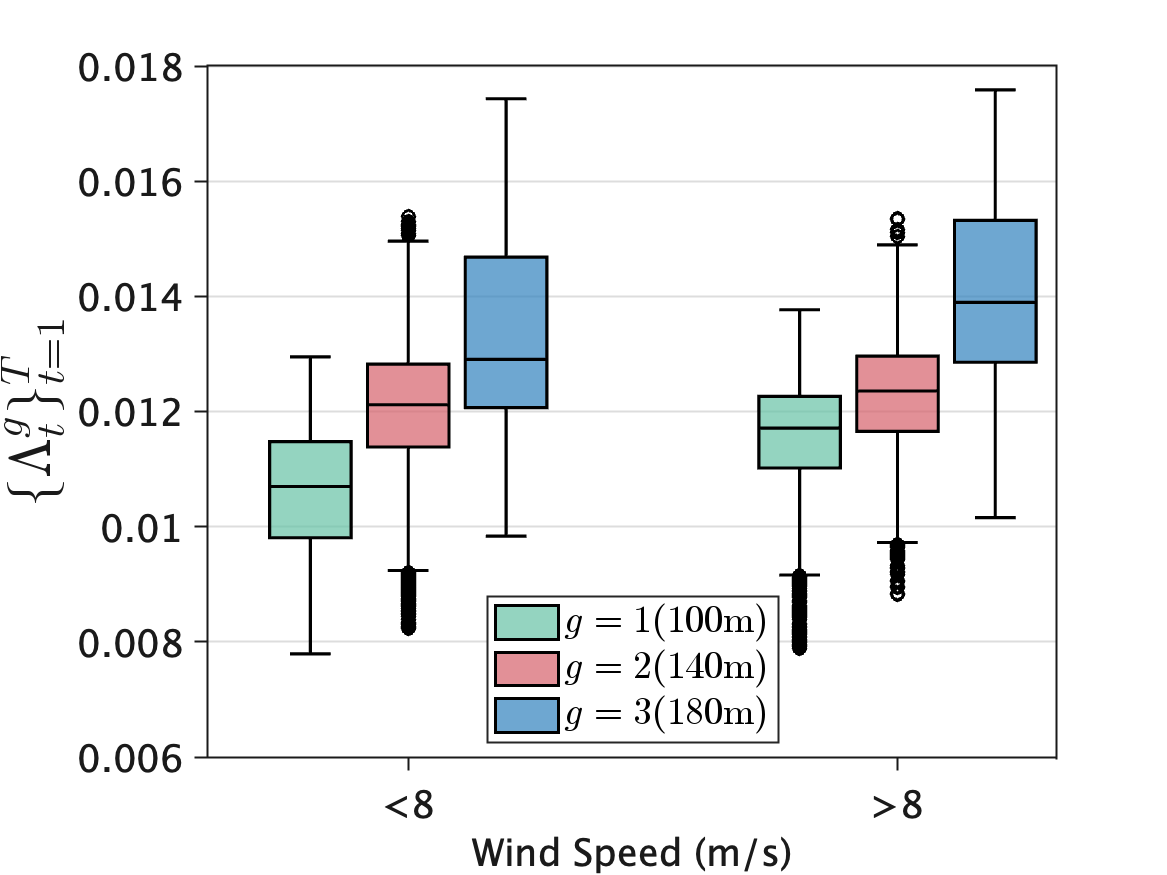}
\caption{Boxplot showing the norm of the predicted advection vectors $\{\Lambda_t^g\}_{t=1}^T$. for different heights and wind speeds.}
\label{fig: adv_bx}  
\end{figure}

To further elucidate the inner workings of DeepMIDE, we take a closer look at the advection parameter estimates, which are produced using the deep-learning-based functional mapping. Let the norm of the estimated advection vectors at height $g$ and time $t$ denote as $\Lambda_t^g=\|\boldsymbol{\theta}_t^g\|$.
Figure \ref{fig: adv_map} shows the time series plot of $\Lambda_t^2$ at height $140$m (corresponding to height index $g = 2$ in our study), clearly illustrating a dynamic pattern over time. We zoom in on two consecutive NWP wind speed maps from 21-Jun-2021 02:00:00 to 03:00:00 (orange rectangle, left) and from 04-Jul-2021 05:00:00 to 06:00:00 (green rectangle, right). During the first time interval, the wind speed maps visually show strong advection effects over time, which appears to be captured by the model as reflected by the relatively large advection parameter predictions in the corresponding time series. In contrast, the second time interval shows significantly weaker advection effects (with the two consecutive images being almost static), with correspondingly smaller predicted values in the advection parameter time series. These two examples demonstrate how the deep learning model is able to translate the information from those maps into physically meaningful estimates of advection. Figure \ref{fig: adv_bx} shows the boxplots of all $\{\Lambda_t^g\}_{t=1}^T$, partitioned by wind speed (x-axis) and height (different colors), which clearly shows that the values of the predicted advection parameters increase at greater heights and stronger winds\textemdash a conclusion which we have observed in our preliminary data analysis of Section \ref{asy}. This further confirms the ability of DeepMIDE to {learn key physical insights of the wind field observed from the data.} 


\section{Conclusions}\label{conclusion}
As the upscaling of offshore wind turbines to unprecedented heights continues, multi-height wind speed forecasts across the now-much-larger rotor area of those ultra-scale generators are  needed. In this work, we propose the first attempt for multi-height spatio-temporal wind energy forecasting. Our model, dubbed DeepMIDE, is formulated as a {multi-output} integro-difference equation model, which fully embraces the full vertical wind profile by incorporating spatial, temporal, and height dependencies, in order to reflect the true complexity of wind interactions coincident with ultra-scale offshore wind turbines. An embedded deep learning architecture directly acts on streams of high-dimensional exogenous weather information in order to encode the physics of local wind field formation in a low-dimensional set of information-rich advection vectors. Those vectors, along with other parameters, are then plugged back into the statistical model for probabilistic space-time forecasting. The proposed approach marries the probabilistic rigor and parsimony of the {multi-output} integro-difference equation framework, with the representative power of the advanced deep transformer architecture which are used as a ``physics extractor'' within the statistical model. Evaluated using real-world data from the {Northeastern} U.S., our model demonstrates significant improvements in both wind speed and power forecasting compared with prevalent forecasting benchmarks.

Future research will investigate more sophisticated modeling structures for the latent process model within the {multi-output integro-difference equation} framework. This may include handling non-Gaussian data and non-linear dependence, which would obviate the need for data transformations or linear assumptions, but would likely complicate the parameter inference process. 
{Additionally, DeepMIDE could also be extended to model bivariate wind vectors rather than directly acting on wind speed time series. Modeling the wind vector is advantageous because directional transport is central to advection modeling across space and time. It is also beneficial in practice for characterizing wind turbine performance. 
Methodologically, this requires fundamentally different statistical treatment to jointly model the wind vector components, which are inherently dependent and directionally structured, through a suitable bivariate distribution.} 


\section*{Online Supplementary Materials}
The supplementary materials appended to this manuscript include: {(1) Details of the Box-Cox transformation}; (2) Details of the architecture and implementation of the deep learning model shown in Figure \ref{fig-dl-model}; and (3) Access to codes and data used to reproduce the results of DeepMIDE.   

\section*{Acknowledgmenets} This work is supported in part by the U.S. National Science Foundation (ECCS-2114422). 

\section*{Disclosure Statement} The authors report there are no competing interests to declare.

\bibliographystyle{jasa3}
\bibliography{references}       

\end{document}